\pdfoutput=1
\documentclass[english]{article}
\usepackage{iclr2015,times}
\usepackage{amstext,amsmath}
\usepackage{url}
\usepackage{graphicx}
\usepackage{multirow}
\graphicspath{ {figures/pdf/}{figures/jpeg/}{figures/png/} {figures/eps/} }
\DeclareGraphicsExtensions{.pdf,.eps,.jpeg,.JPEG,.jpg,.png}
\usepackage{caption}
\usepackage{subcaption}
\usepackage{pbox}
\usepackage{tabularx}

\usepackage[compact]{titlesec}
\titlespacing{\section}{0pt}{*1}{*0}
\titlespacing{\subsection}{0pt}{*0}{*0}
\titlespacing{\subsubsection}{1pt}{*1}{*0}

\newcommand*{\widthfactor}{0.16}
\newcommand*{\imagewidth}{0.9in}

\makeatother

\usepackage{babel}
\title{Understanding Locally Competitive\\ Networks}

\author{
    Rupesh Kumar Srivastava,
    Jonathan Masci,
    Faustino Gomez \&
    J{\"u}rgen Schmidhuber\\
    Istituto Dalle Molle di studi sull'Intelligenza Artificiale (IDSIA)\\
    Scuola universitaria professionale della Svizzera italiana (SUPSI)\\
    Universit\`a della Svizzera italiana (USI)\\
    Lugano, Switzerland\\
{\{\tt rupesh,\! jonathan,\! tino,\! juergen\}@idsia.ch}\\
    }

\begin{document}

\iclrconference
\iclrfinalcopy

\maketitle

\begin{abstract} 
Recently proposed neural network activation functions such as
rectified linear, maxout, and local winner-take-all have allowed for
faster and more effective training of deep neural architectures on
large and complex datasets.  The common trait among these
functions is that they implement local competition between small
groups of computational units within a layer, so that only part of the network
is activated for any given input pattern. 
In this paper, we attempt to visualize and understand this self-modularization, 
and suggest a unified explanation for the beneficial properties
of such networks. We also show how our insights can be 
directly useful for efficiently performing retrieval over large datasets
using neural networks.
\end{abstract}

\section{Introduction}
Recently proposed activation functions for
neural networks such as rectified linear (ReL~\citep{glorot2011}), 
maxout \citep{goodfellow2013a} and LWTA \citep{srivastava2013a}
are quite unlike sigmoidal activation functions.
These functions depart from the conventional wisdom in that they are not
continuously differentiable (and sometimes non-continuous) and are
piecewise linear.  Nevertheless, many researchers have found that such
networks can be trained faster and better than sigmoidal networks, and
they are increasingly in use for learning from large and complex
datasets \citep{krizhevsky2012,zeiler2013a}. Past research has shown
observational evidence that such networks have beneficial 
properties such as not requiring unsupervised training for weight
initialization \citep{glorot2011}, better gradient flow \citep{goodfellow2013a}
and mitigation of catastrophic forgetting \citep{srivastava2013a,goodfellow2014}.
Recently, the expressive power of deep networks with such functions
has been theoretically analyzed \citep{pascanu2013}.
However, we are far from a complete understanding of their
behavior and advantages over sigmoidal networks,
especially during learning.
This paper sheds additional light
on the properties of such networks by interpreting
them as \emph{models of models}.

A common theme among the ReL, maxout and LWTA
activation functions is that they are
locally competitive. Maxout and LWTA utilize explicit competition
between units in small groups within a layer, while in the case of the
rectified linear function, the weighted input sum competes with a
fixed value of 0. Networks with such functions are often trained with
the dropout regularization technique \citep{hinton2012} for improved
generalization. 

We start from the observation that in locally competitive networks,
a subnetwork of units has non-zero activations for
each input pattern. 
Instead of treating a neural network as a complex function approximator,
the expressive power of the network can be
interpreted to be coming from its ability to activate
different subsets of linear units for different patterns.
We hypothesize that the network acts as a model that can
switch between ``submodels'' (subnetworks) 
such that \emph{similar submodels respond to similar patterns}. 
As evidence of this behavior, 
we analyze the activated subnetworks for a large
subset of a dataset (which is not used for training) 
and show that the subnetworks activated for
different examples exhibit a structure consistent with our hypothesis.
These observations provide a unified explanation for improved credit assignment
in locally competitive networks during training, 
which is believed to be the main reason for their success.
Our new point of view suggests a link between these networks
and competitive learning approaches of the past decades.
We also show that a simple encoding of which units in a layer are 
activated for a given example (its subnetwork) can be used 
to represent the example for retrieval tasks. 
Experiments on MNIST, CIFAR-10, CIFAR-100 and the 
ImageNet dataset show that promising results are obtained
for datasets of varying size and complexity.

\begin{figure} 
  \centering
  \begin{minipage}{0.48\textwidth}
  \centering
  \includegraphics[scale=0.2]{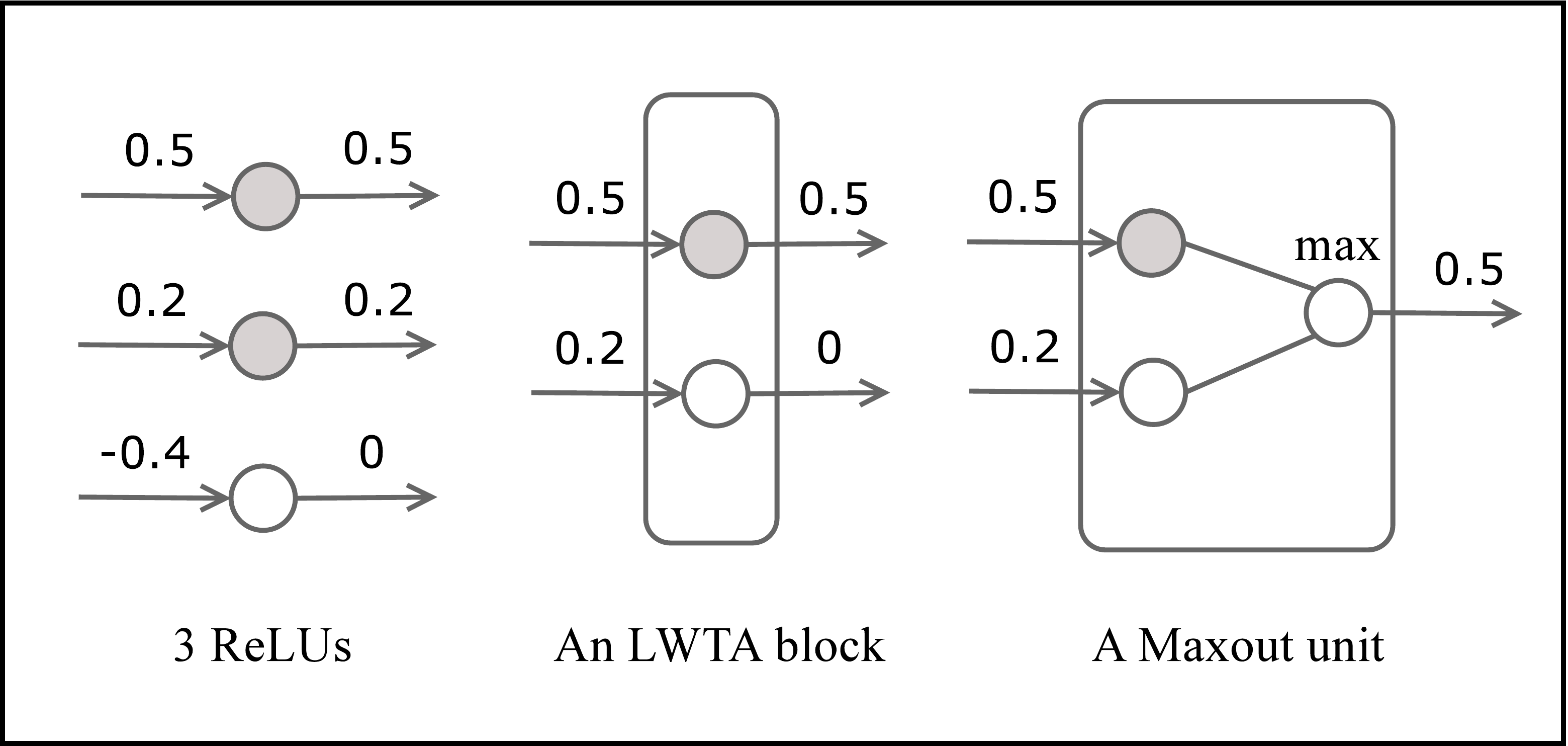}
  \caption{Comparison of rectified linear units (ReLUs), local winner-take-all (LWTA), and 
  maxout activation functions. The pre- and post-synaptic activations of the units are
  shown on the left and right side of the units respectively. The shaded units are `active' -- 
  non-zero activations and errors flow through them. The main difference
  between maxout and LWTA is that the post-synaptic activation can flow through
  connections with different weight depending on the winning unit in LWTA. For
  maxout, the outgoing weight is the same for all units in a block.}
  \label{fig:actfunctions}
  \end{minipage}
  \hfill
  \begin{minipage}{0.48\textwidth}
  \centering
  \includegraphics[scale=0.35]{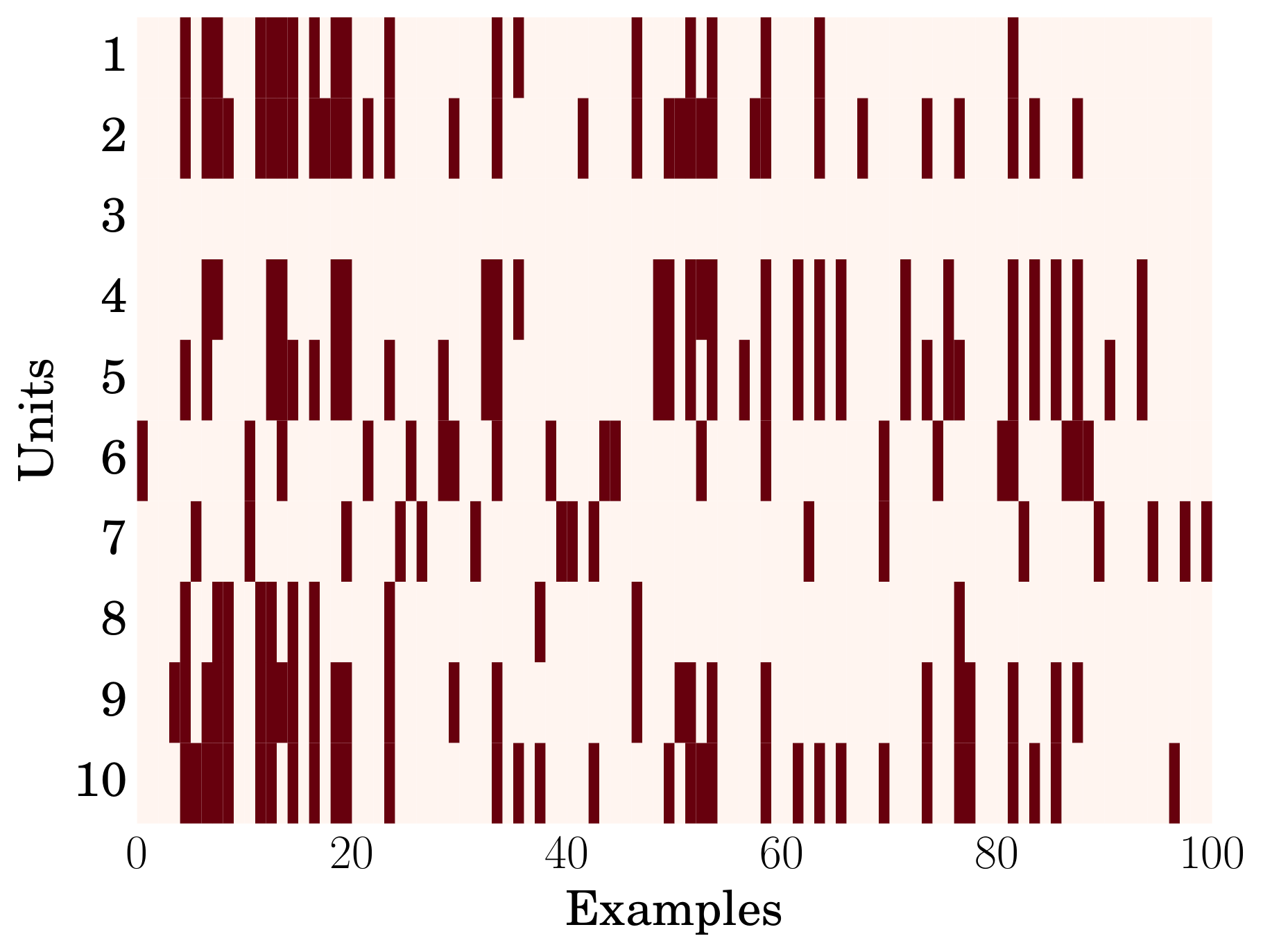}
  \caption{Subnetworks for 100 examples for 10 ReLUs. The examples activate many different
  possible subsets of the units, shown in dark. In this case, unit number 3 is inactive for all examples.}
  \label{fig:subnets}
  \end{minipage}
\end{figure}

\section{Locally Competitive Neural Networks} 
Neural networks with activation functions like rectified linear, maxout and
LWTA are locally competitive. This means that 
local competition among units in the network decides
which parts of it get
activated or trained for a particular input example.
For each unit, the total input or
presynaptic activation $z$ is first computed as $z = \mathbf{wx} + b$, where $\mathbf{x}$ is the
vector of inputs to the unit, $\mathbf{w}$ is a trainable weight vector, and $b$ is a
trainable bias.  For the rectified linear function, the output or postsynaptic
activation of each unit is simply $max(z, 0)$, which can be interpreted as
competition with a fixed value of 0. For LWTA, the units in a layer are
considered to be divided into blocks of a fixed size. Then the output of each
unit is $Iz$ where $I$ is an indicator which is 1 if the unit has the
maximum $z$ in its group and 0 otherwise. In maxout, the inputs from a few
units compete using a $max$ operation, and the block output is the maximum $z$ among
the units\footnote{In our terminology, the terms \emph{unit} and \emph{block} 
correspond to the terms \emph{filter} and \emph{units} in \citet{goodfellow2013a}.}. 
A maxout block can also be interpreted as an LWTA block with shared
outgoing weights among the units. A comparison of the 3 activation functions
is shown in Figure \ref{fig:actfunctions}. 

In each of the three cases, there is a local gating mechanism which
allows non-zero activations (and errors during training) to propagate only
through part of the network, i.e.\ a subnetwork.  Consider the activation of
a neural network with rectified linear units (ReLUs) in a single hidden layer.  For each input
pattern, the subset of units with non-zero activations in the hidden
layer form a subnetwork, and an examination of the
subnetworks activated for several examples shows that a large number
of different subnetworks are activated (Figure \ref{fig:subnets}).
The result of training the network can interpreted in the following
way: when training a single network with a local gating mechanism, a
large number of linear subnetworks are trained on the dataset such
that different examples are gated to different subnetworks, each
getting trained to produce the desired output. At test time, the
system generalizes in the sense that the appropriate subnetwork for a
given example is activated.

\section{Subnetwork Analysis}
\label{sec:subnetwork-analysis}

\begin{figure} 
  \centering 
  \begin{subfigure}[t]{0.48\textwidth}
  \centering 
    \includegraphics[scale=0.4]{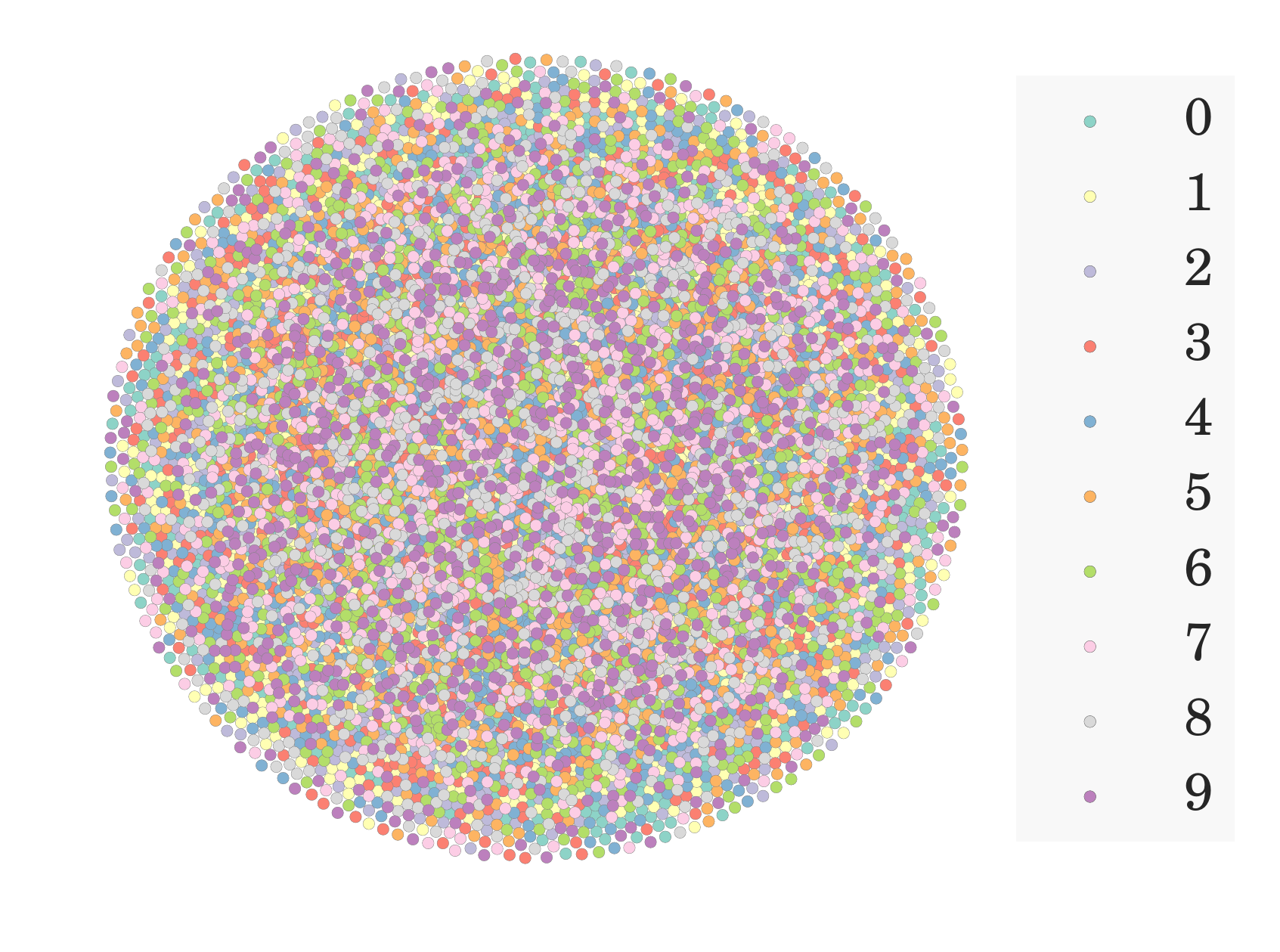} 
    \caption{}
    \label{fig:tsne-layer3-mnist-relu-untrained} 
  \end{subfigure} 
  \hfill
  \begin{subfigure}[t]{0.48\textwidth}
  \centering 
    \includegraphics[scale=0.4]{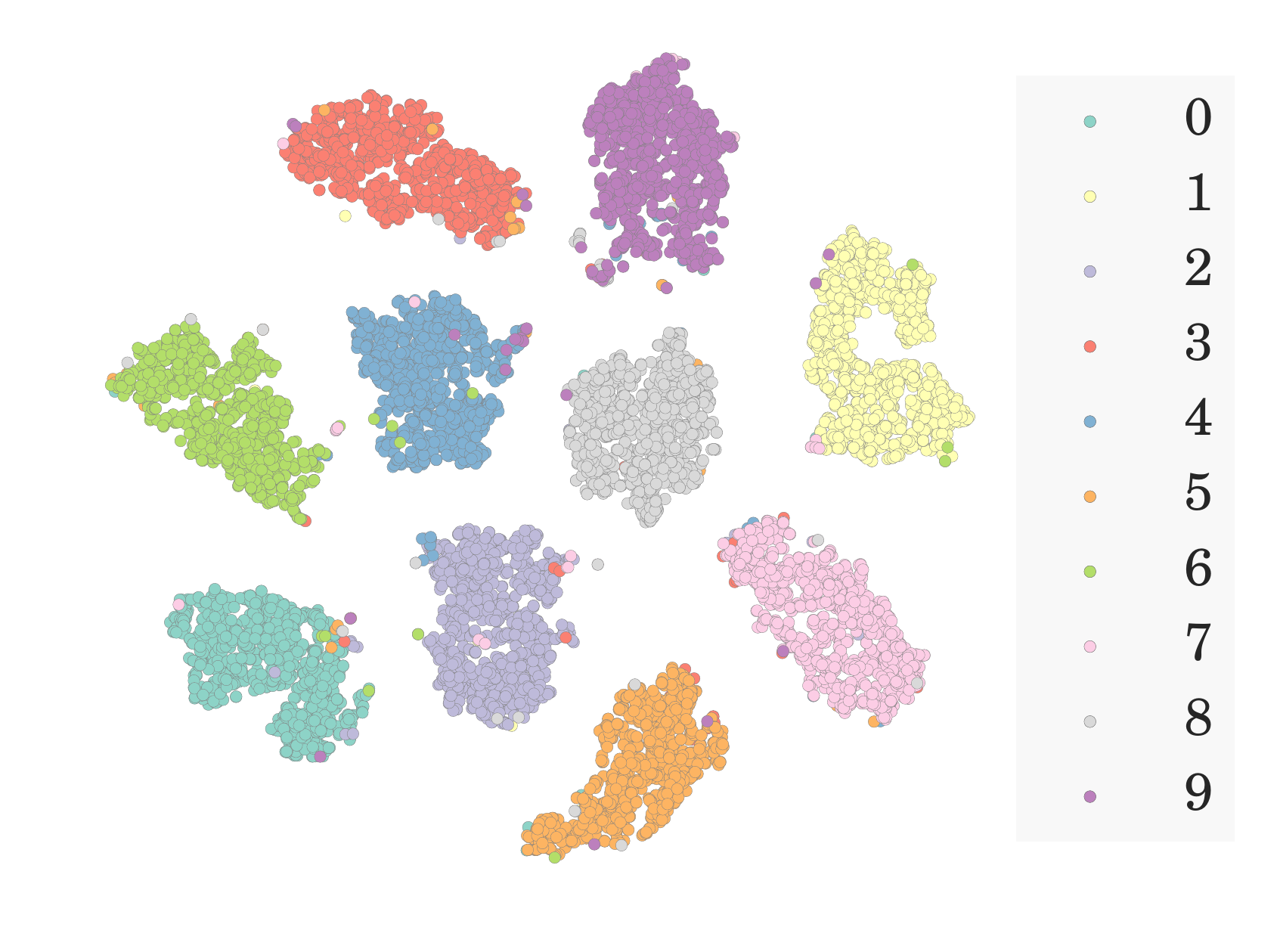}
    \caption{}
    \label{fig:tsne-layer3-mnist-relu-trained} 
  \end{subfigure} 
  \caption{2-D visualization of submasks from the penultimate layer of
    a 3 hidden layer network with ReLUs on the MNIST test set. (a)
    shows the submasks from an \emph{untrained} network layer which
    lacks any discernable structure.  (b) shows submasks from a
    trained network layer, showing clearly  demarcated clusters relevant to the
    supervised learning task. `Mistakes' made by the network can also
    be observed, such as mistaking `4's for `9's.}
\end{figure}

This section investigates how the model of models that is implemented
though local competition self-organizes due to training. 
In order to visualize the organization of subnetworks as a result of
training, they are encoded as bit strings called
\emph{submasks}.  For the input pattern $i$, the submask $s_i\in
\{0,1\}^u$, where $u$ is the number of units in the full network,
represents the corresponding subnetwork by having a 0 in position $j$,
$j=1..u$, if the corresponding unit has zero activation, and 1
otherwise.  The submasks uniquely and compactly encode each subnetwork
in a format that is amenable to analysis through clustering, and, as we
show in Section~\ref{sec:imagenet}, facilitates efficient data retrieval.

In what follows, the subnetworks that emerge during training are
first visualized using the t-SNE~\citep{vandermaaten2008} algorithm. 
This dimensionality reduction technique enables a good visualization 
of the relationship between submasks for
several examples in a dataset by preserving the local structure. 
Later in this section, we examine the evolution of subnetworks during training, 
and show that the submasks obtained from a trained 
network can directly be used for classification using 
a simple nearest neighbors approach.
All experiments in this section are performed on the 
MNIST~\citep{lecun1998} dataset.  This familiar dataset was chosen because it
is relatively easy, and therefore provides a tractable setting
in which to verify the repeatability of our results.
Larger, more interesting datasets are used in section~\ref{sec:exp} to
demonstrate the utility of techniques developed in this section for
classification and retrieval.

\subsection{Visualization through Dimensionality Reduction}
For visualizing the relationship between submasks for a large number
of input patterns, we trained multiple networks with different
activation functions on the MNIST training set, stopping when the
error on a validation set did not improve.  The submasks for the
entire test set (10K examples) were then extracted and visualized
using t-SNE.  Since the competition between subnetworks is local and
not global, subsets of units in deeper (closer to the output) layers
are activated based on information extracted in the shallow
layers. Therefore, like unit activations, submasks from deeper layers
are expected to be better related to the task since deeper layers code
for higher level abstractions.  For this reason, we use only submasks
extracted from the penultimate network layers in this paper, which
considerably reduces the size of submasks to consider.

Figure~\ref{fig:tsne-layer3-mnist-relu-trained} shows a 2D
visualization of the submasks from a 3 hidden layer ReL network. Each
submask is a bitstring of length 1000 (the size of the network's
penultimate layer).  Ten distinct clusters are present corresponding
to the ten MNIST classes. 
It is remarkable that, irrespective of the actual activation 
values, the subnetworks which are active for
the testing examples can be used to visually predict class
memberships based on their similarity to each other. 
The visualization confirms that the subnetworks
active for examples of the same class are much more similar to each
other compared to the ones activated for the examples of different
classes. 

Visualization of submasks from the same layer of a 
randomly initialized network does not
show any structure
(Figure~\ref{fig:tsne-layer3-mnist-relu-untrained}), but we
observed some structure for the untrained first hidden layer 
(Appendix \ref{sec:more-vis}).
For trained networks, similar clustering is 
observed in the submasks from shallow layers in the network, though
the clusters appear to be less separated and tight.  The visualization
also shows many instances where the network makes mistakes. The
submasks for some examples lie in the cluster of submasks for the
wrong class, indicating that the `wrong' subnetwork was selected for
these examples.
The experiments in the next sections show that the organization of
subnetworks is indicative of the classification performance of the
full network.

Other locally competitive activation functions such as LWTA and maxout
result in similar clustering of submasks (visualizations 
included in Appendix \ref{sec:more-vis}). For LWTA layers, the submasks can
be directly constructed from the activations because there is no
subsampling when going from presynaptic to postsynaptic activations,
and it is reasonable to expect a subnetwork organization similar to 
that of ReL layers. 
Indeed, in a limited qualitative analysis, it has been shown previously 
\citep{srivastava2013a} that in trained LWTA nets there
are more units in common between subnetworks for examples of the same class
than those for different class examples.

For maxout layers, the situation is trickier at a first glance. The unit
activations get pooled before being propagated to the next layer, so it is
possible that the maximum activation value plays a much more important role
than the identity of the winning units. However, using the same basic principle
of credit assignment to subnetworks, we can construct submasks from maxout layers
by binarizing the unit activations such that only the units producing the
maximum activation are represented by a 1. Separation of subnetworks 
is necessary to gain the advantages of local competition during learning, 
and the visualization of the generated submasks produces results similar to those
for ReLU and LWTA (included in Appendix \ref{sec:more-vis}).

\subsection{Behavior during Training}
\begin{figure} 
  \begin{minipage}{0.53\textwidth}
  \centering
  \includegraphics[scale=0.4]{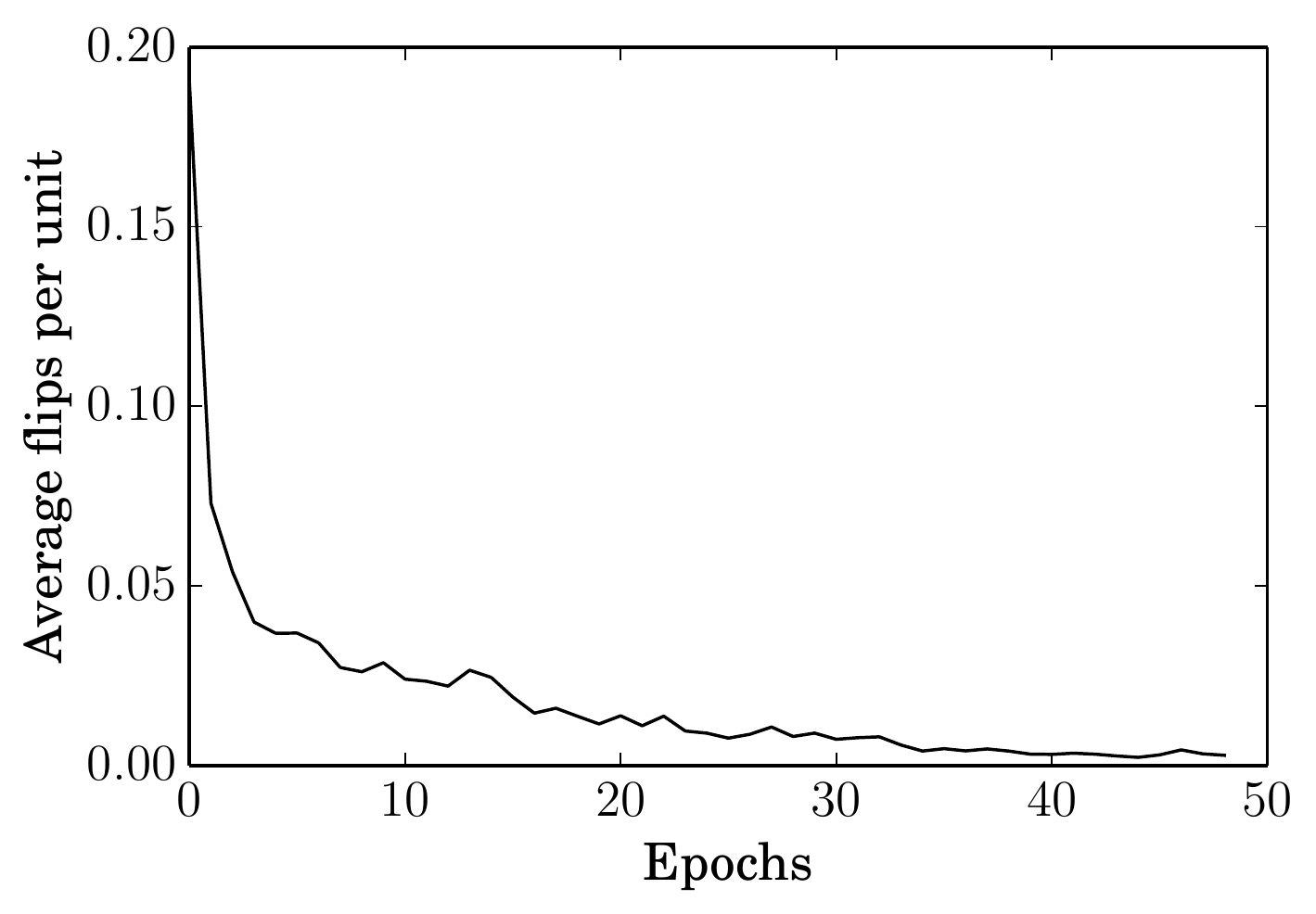} 
  \caption{The plot shows
    mean of the fraction of examples (total 10K) for which units in the 
    layer flip (turn from being
    active to inactive or vice-versa) after every pass through the training set.
    The units flip for upto 20\% of the examples on average in the first few
    epochs, but quickly settle down to less than 5\%. }
  \label{fig:relu-mean-over-time}
  \end{minipage}
  \hfill
  \begin{minipage}{0.45\textwidth}
  \centering
  \begin{tabular*}{\textwidth}{lcc}
    \hline
    \multirow{2}{*}{Network}      &\multicolumn{2}{c}{No. test errors}\\
                                  & Softmax        & $k$NN      \\ \hline
    ReL (no dropout)     & 161            & 158          \\
    LWTA (dropout)        & 142              & 154          \\
    Maxout (dropout)      & 116            & 131          \\ \hline
    \end{tabular*}
    \captionof{table}{Some examples of classification results on the 
    permutation invariant MNIST test set using 
    softmax layer outputs vs. $k$NN on the submasks. All
    submasks are extracted from the penultimate layer.
    $k$NN results are close to the softmax results in each
    case. The maxout network was additionally 
    trained on the validation set. Results vary slightly 
    across experimental runs and were not cherry-picked for reporting.}
  \label{tab:knn-vs-softmax}
  \end{minipage}
\end{figure}
In order to measure how the subnetworks evolve over the course of
training, the submasks of each sample in the training set were
recorded at each epoch.  Figure~\ref{fig:relu-mean-over-time}
characterizes the change in the subnets over time by counting the
number of input patterns for which a unit flips from being on to being
off, or vice-versa, from one epoch to the next.  The curve in the
figure shows the fraction of patterns for which an inter-epoch flip
occurred, averaged across all units in the network.  Higher values
indicate that the assignment of subnets to patterns is not stable.
The batch size for this experiment was 100, which means that each pass
over the training set consists of 500 weight updates. For the run
shown, the average fraction of flips starts at 0.2, but falls quickly
below 0.05 and keeps falling as training proceeds, indicating that,
the assignment of subnetworks to individual examples stabilizes quickly. 
In this case, after a brief ($\sim$3 epochs) transient period, a fine-tuning 
period follows where the assigned subnetworks keep getting 
trained on their corresponding examples with little re-assignment.

\subsection{Evaluating Submasks}
Since the visualization of submasks for the test set shows
task-relevant structure, it is natural to ask: how well can the submask
represent the data that produced it? 
If the submasks for similar examples are similar, perhaps they can
be used as data descriptors for tasks such as similarity-based retrieval.
Sparse binary codes enable efficient
storage and retrieval for large and complex datasets due to which
learning to produce them is an active research area 
\citep{gong2013,masci2014,masci2014pami,grauman2013}.
This would make representative submasks very attractive since no 
explicit training for retrieval would be required to generate them.

To evaluate if examples producing similar binary codes are indeed 
similar, we train locally competitive networks for classification and  
use a simple $k$ nearest neighbors ($k$NN) algorithm 
for classifying data using the generated submasks.
This approach is a simple way to examine the amount of information
contained in the submasks (without utilizing the actual activation
values).

We trained networks with fully connected layers on
the MNIST training set, and selected the value 
of $k$ with the lowest validation
error to perform classification on the test set. 
Results are shown in Table~\ref{tab:knn-vs-softmax}. In each case, the
$k$NN classification results are close to the classification result
obtained using the network's softmax layer. If we use the 
(non-pooled) unit activations from the maxout network instead of 
submasks for $k$NN classification, we obtain 121 errors.

Submasks can also be obtained from convolutional layers.
Using a convolutional maxout network, we obtained 52 errors on the
MNIST test set when we reproduced the model from
\citet{goodfellow2013a}. Since the penultimate layer in this model is
convolutional, the submasks were constructed
using the presynaptic unit activations from this layer for all
convolutional maps. Visualization of these submasks showed similar
structure to that obtained from fully connected layers, $k$NN
classification on the submasks resulted in 65 errors. As seen before,
for a well-trained network the $k$NN performance is close to the
performance of the network's softmax layer.

\subsection{Effect of Dropout} 
The dropout~\citep{hinton2012} regularization technique has proven to be very
useful and efficient at improving generalization for large models, and
is often used in combination with locally competitive activation
functions~\citep{krizhevsky2012,goodfellow2013a,zeiler2013a}.  We found
that networks which were trained with dropout (and thus produced lower
test set error) also yielded better submasks in terms of $k$NN
classification performance. To observe the effect of dropout in more
detail, we trained a 3 hidden layer network with 800 ReLUs in each
hidden layer without dropout on MNIST starting from 5
different initializations until the validation set error did not
improve. The networks were then trained again from the same
initialization with dropout until the validation error matched or fell
below the lowest validation error from the non-dropout case.  In both
cases, minibatch gradient descent with momentum was used for training the
networks.  A comparison of $k$NN classification error for the dropout
and non-dropout cases showed that when dropout training is stopped 
at a point when validation error is similar to a no-dropout network, 
the submasks from both cases give similar results, but as
dropout training continues (lowers validation set error), the
submasks yield improved results. 
This supports the interpretation of dropout as a regularization technique 
which prevents ``co-adaptation of feature detectors'' (units)~\citep{hinton2012}, 
leading to better representation of data by the subnetworks. 
Another way to look at this effect can be that dropout improves generalization
by injecting noise in the organization of subnetworks, making them more robust.

\begin{figure}
  \centering
  \begin{subfigure}[t]{0.48\textwidth}
  \centering 
    \includegraphics[scale=0.4]{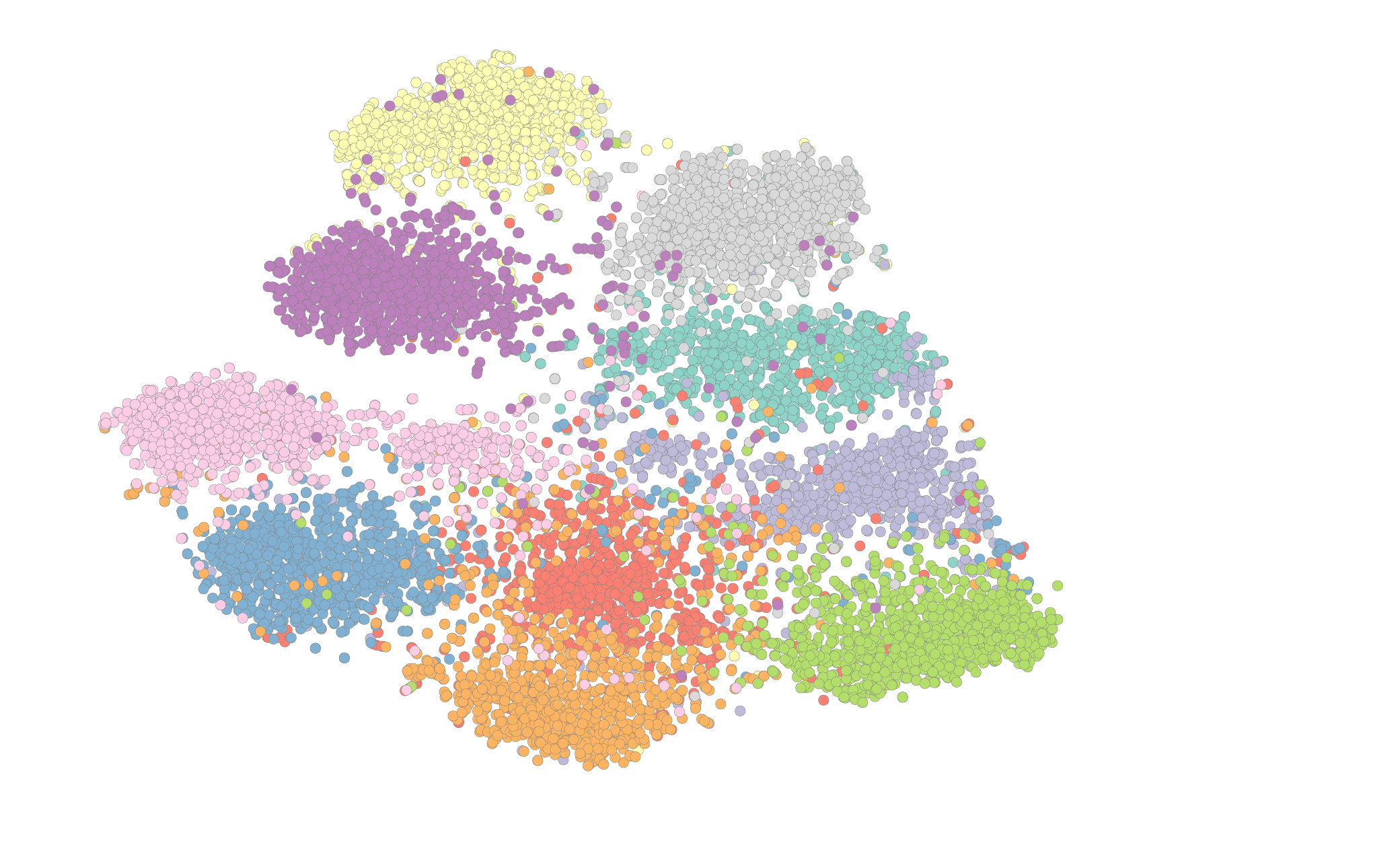} 
    \caption{} 
    \label{fig:tsne-layer3-cifar10-maxout} 
  \end{subfigure}
  \hspace{10pt}
  \begin{subfigure}[t]{0.48\textwidth}
  \centering 
    \includegraphics[scale=0.4]{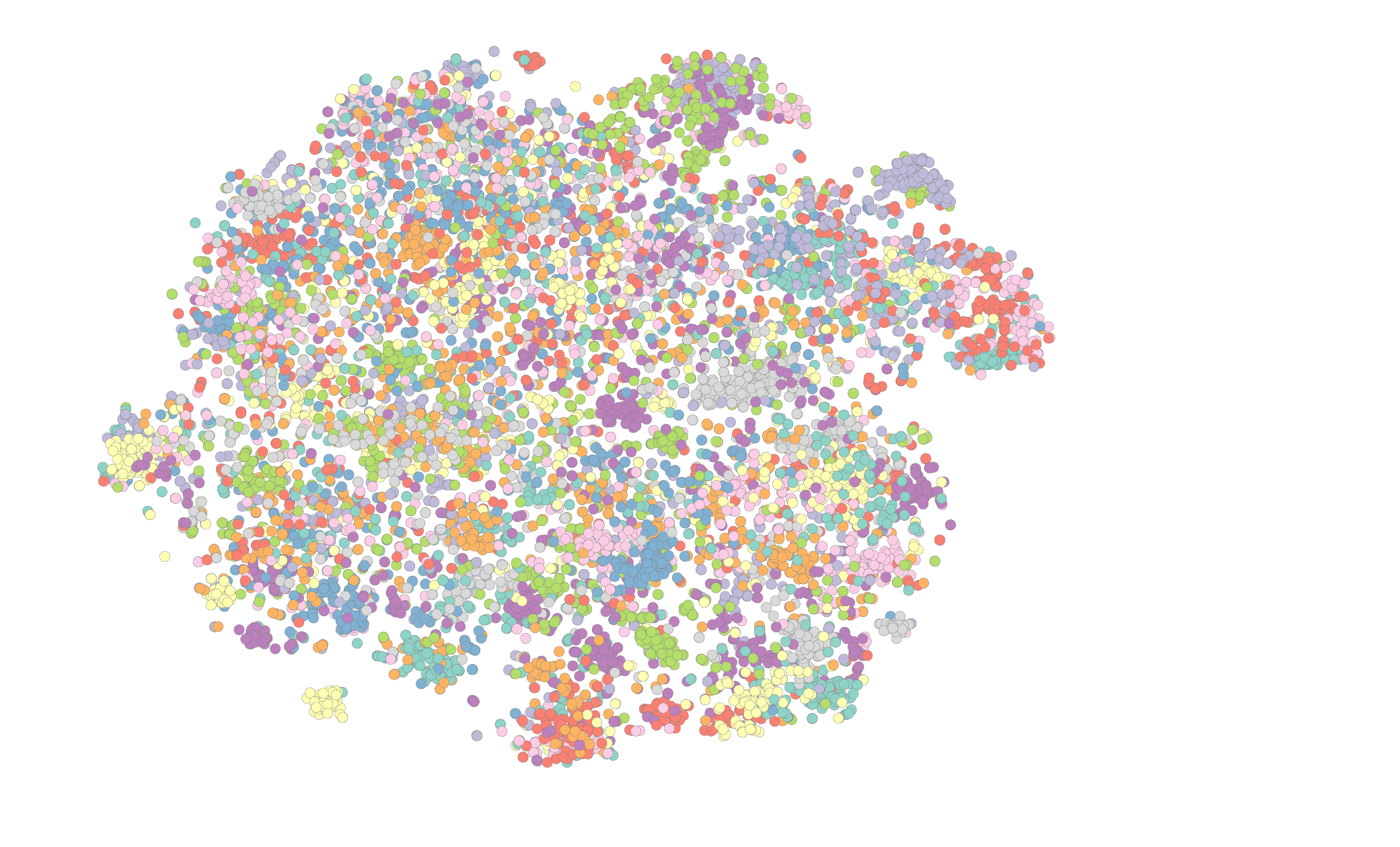} 
    \caption{} 
    \label{fig:tsne-layer3-cifar100-maxout} 
  \end{subfigure} 
  \caption{2-D visualizations of the submasks from the penultimate 
    layer of the trained maxout networks reported in \citet{goodfellow2013a}. 
    (a) The CIFAR10 test set. The 10-cluster structure is visible,
     although the clusters are not as well separated as in the case of MNIST.
    This corresponds with the higher error rates obtained using both $k$NN 
    and the full network. (b) The CIFAR100 test set.
    It is difficult to visualize any dataset 
    with 100 classes, but several clusters are still visible. 
    The separation between clusters is much worse, which 
    is reflected in the high classification error.}
\end{figure}

\section{Experimental Results} 
\label{sec:exp}
The following experiments apply the methods described in the previous
section to more challenging benchmark problems: CIFAR-10, CIFAR-100,
and ImageNet.  For the CIFAR experiments, we used the models described
in~\citet{goodfellow2013a} since they use locally competitive activations 
(maxout), are trained with dropout, and good hyperparameter settings for 
them are available~\citep{goodfellow2013}.
We report the classification error on the test set obtained using the softmax
output layer, as well $k$NN classification on the penultimate layer
unit activations and submasks. The best value of $k$ is obtained using 
a validation set, though we found that $k$ = 5 with distance weighting 
usually worked well.

\subsection{CIFAR-10 \& CIFAR-100}
CIFAR-10 and CIFAR-100 are datasets of 32$\times$32 color images of 10 classes.
The results obtained on the test sets for these datasets are summarized in Table
\ref{tab:cifar-10-100}.
We find that when comparing nearest neighbor classification performance with 
submasks to unit activation values, we lose an accuracy of 1.25\% on the CIFAR-10 
dataset, and 2.26\% on the CIFAR-100 dataset on average.
Figure \ref{fig:tsne-layer3-cifar10-maxout} shows the 2-D visualization of the test set 
submasks for CIFAR-10. Some classes can be seen to have highly representative 
submasks, while confusion between classes in the lower half is observed. 
The clusters of subnetworks are not as well-separated as in the case of MNIST, 
reflecting the relatively worse classification performance of the full network.
Submask visualization for CIFAR-100 (Figure \ref{fig:tsne-layer3-cifar100-maxout})
reflects the high error rate for this dataset. Although any visualization with 100 classes 
can be hard to interpret, many small clusters of submasks can still be observed.

\begin{table}
  \begin{tabularx}{\textwidth}{XXXXX}
    \hline
    Dataset   & Network error & $k$NN\newline(activations) & $k$NN\newline(pre-activations) & $k$NN\newline(submasks)\\ \hline
    CIFAR-10  & 9.94 $\pm$ 0.31\%  & 9.63 $\pm$ 0.21\%  &  10.11 $\pm$ 0.16\% & 11.36 $\pm$ 0.22\%  \\
    CIFAR-100 & 34.49 $\pm$ 0.22\% & 37.54 $\pm$ 0.14\% &  41.37 $\pm$ 0.26\%  & 43.63 $\pm$ 0.18\%  \\
\end{tabularx}
    \caption{Classification errors on CIFAR datasets comparing maxout network
    performance, $k$NN on activation values, $k$NN on pre-activations (before maximum pooling) and $k$NN on binary
    submasks. Results are reported over 5 runs.}
    \label{tab:cifar-10-100}
\end{table}

\subsection{ImageNet}
\label{sec:imagenet}
The results of $k$NN classification and t-SNE visualization using submasks on
small datasets of varying complexities show that the submasks contain
substantial information relevant for image classification. 
In this section, the utility of the submasks obtained for a large convolutional 
network trained on the ImageNet Large Scale Visual Recognition 
Challenge 2012 (ILSVRC-2012) \citep{deng2012} dataset is evaluated.

\begin{figure}
  \begin{minipage}{0.50\textwidth}
  \centering
  \includegraphics[scale=0.35]{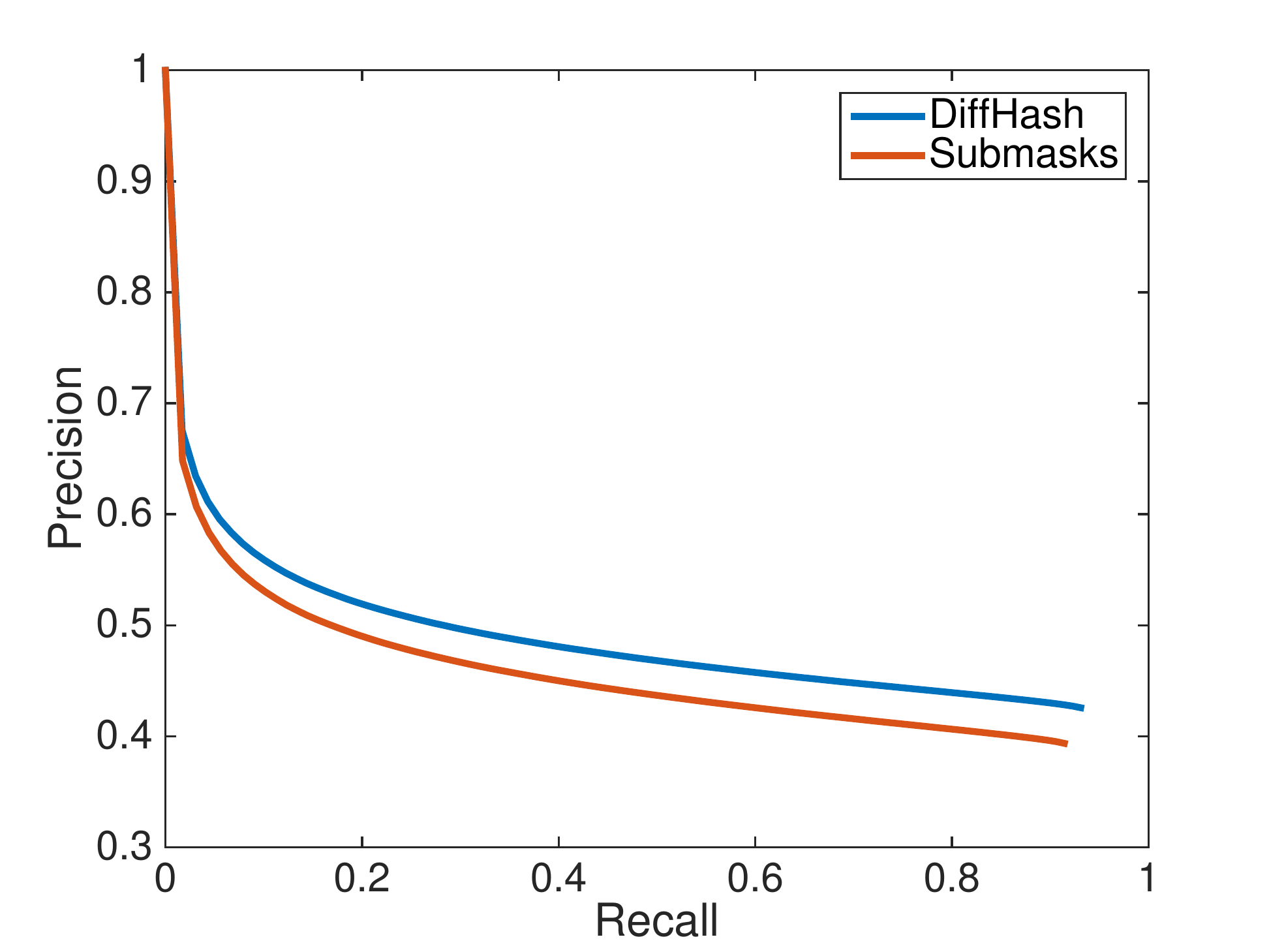}
  \caption{Comparison of precision-recall curves on ILSVRC-2012 when 
  using binary codes obtained using different techniques. 
  The performance of submasks is competitive and decays only for high recall
  values where supervised hashing obtains a better ranking of the results due
  to the pair-wise supervision.
  }
  \label{fig:prcurves}
  \end{minipage}
  \hfill
  \begin{minipage}{0.48\textwidth}
  \small
  \centering
  \begin{tabular*}{\textwidth}{lcc}
    \hline
    Network   & Network error & $k$NN on submasks \\ \hline
    DeCAF     &  19.2\%          &  29.2\%\\
    Convnet\footnotemark\     &  13.5\%          & 20.38\%\\
    \hline
  \end{tabular*}
  \captionof{table}{\small Top-5 Classification accuracy on validation set
  when performance of two different networks on ImageNet is compared to 
  performance of submasks obtained from each of them. Note that as 
  network accuracy improves by about 6\%, submask accuracy improves by about 10\%.}
  \label{tab:imagenet-classification}
  \vspace{10pt}
  \begin{tabular*}{\textwidth}{lccc}
    \hline
    Technique    &  mAP@5      & mAP@10       & mAP@100   \\ \hline
    Submasks     &   58.3      &   56.7      & 46.7     \\
    Diffhash     &   61.0      &   59.3      & 49.5     \\
  \end{tabular*}
    \captionof{table}{\small Comparison of mean average precisions at 
    various thresholds using binary codes obtained using different techniques
on the ILSVRC-2012 dataset. Submasks are obtained directly from networks trained
    for classification without any further training.
    Up to mAP@100 the submasks show a constant performance degradation of about 3 points.
}
  \label{tab:retrieval-table}
  \end{minipage}
  \end{figure}
\footnotetext{https://github.com/torontodeeplearning/convnet/}

Our results show that submasks retain a large amount of information on
this difficult large scale task, while greatly improving storage efficiency.
For instance, 4096-dimensional submasks for the full ILSVRC-2012 training 
set can be stored in about 0.5 GB. Our experiments also indicate that 
submasks obtained from a better trained network result in better performance
(Table \ref{tab:imagenet-classification}).
\citet{krizhevsky2012} suggested that the activations from a trained
convolutional network can be compressed to binary codes using auto-encoders. 
We show here that the submasks can be directly utilized for efficient 
retrieval of data based on high level similarity even though no pair-wise 
loss was used during training.

\begin{figure}[t]
  \centering
  \begin{subfigure}[t]{\widthfactor\textwidth}
    \includegraphics[width=\imagewidth]{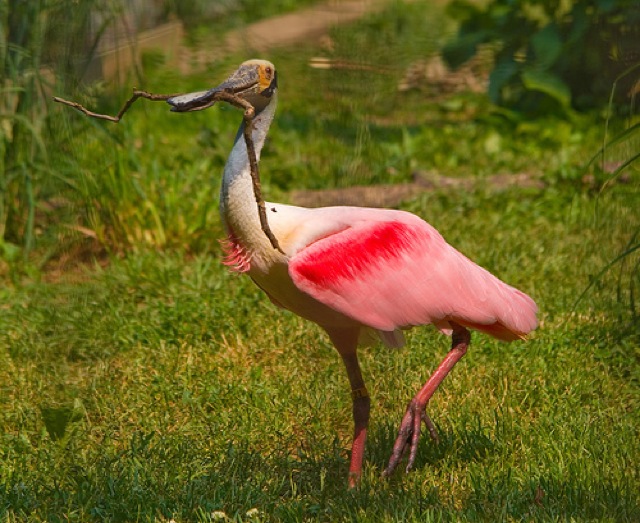}
  \end{subfigure}
  \begin{subfigure}[t]{\widthfactor\textwidth}
    \includegraphics[trim=0 1.3cm 0 1.25cm, clip=True, width=\imagewidth]{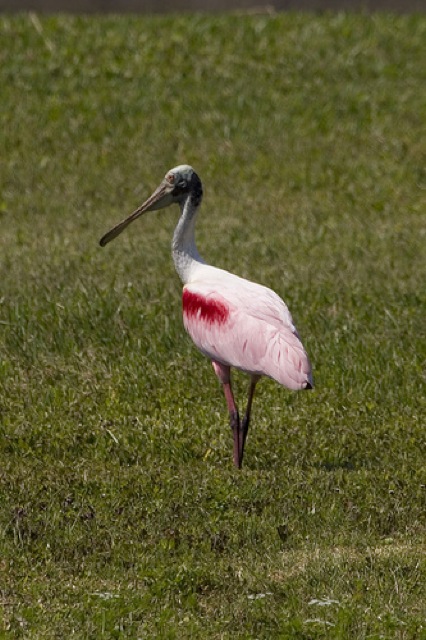}
  \end{subfigure}
  \begin{subfigure}[t]{\widthfactor\textwidth}
    \includegraphics[width=\imagewidth]{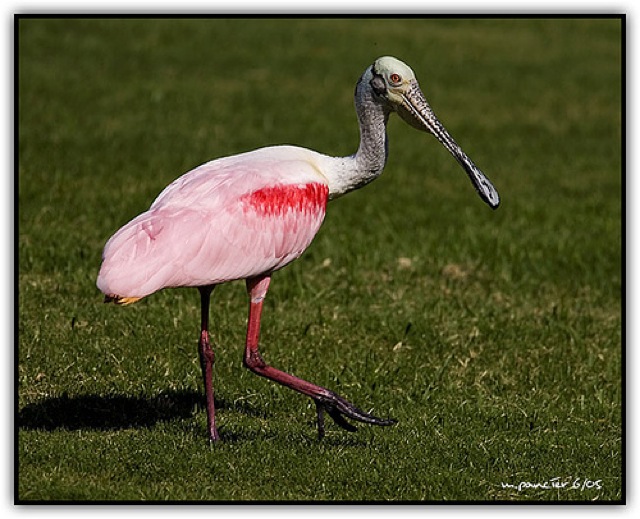}
  \end{subfigure} 
  \begin{subfigure}[t]{\widthfactor\textwidth}
    \includegraphics[width=\imagewidth]{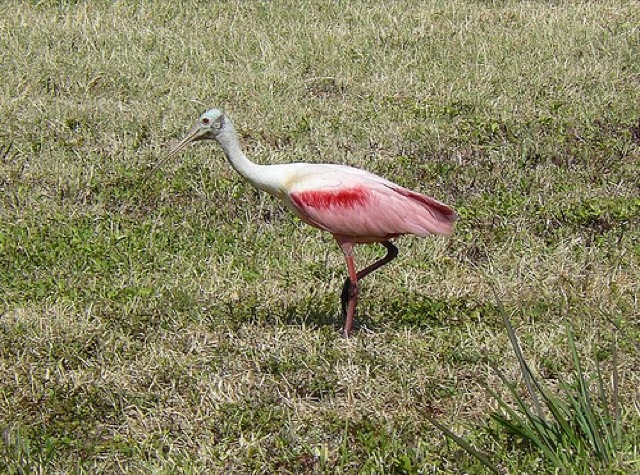}
  \end{subfigure}
  \begin{subfigure}[t]{\widthfactor\textwidth}
    \includegraphics[trim=0 3.0cm 0 4.0cm, clip=True, width=\imagewidth]{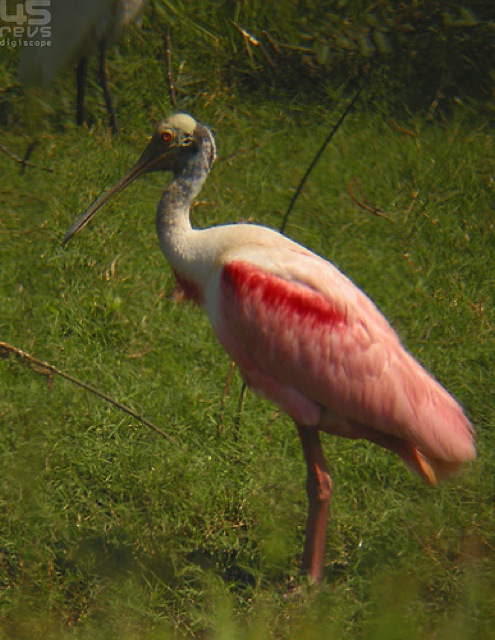}
  \end{subfigure}
  \begin{subfigure}[t]{\widthfactor\textwidth}
    \includegraphics[width=\imagewidth]{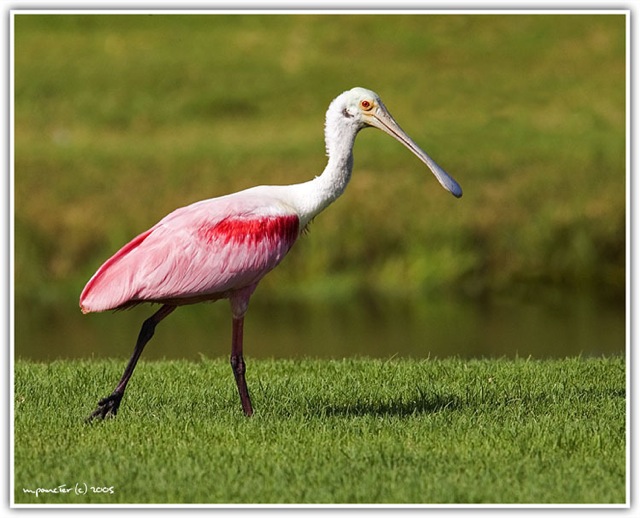}
  \end{subfigure}
  \\
  \begin{subfigure}[t]{\widthfactor\textwidth}
    \includegraphics[width=\imagewidth]{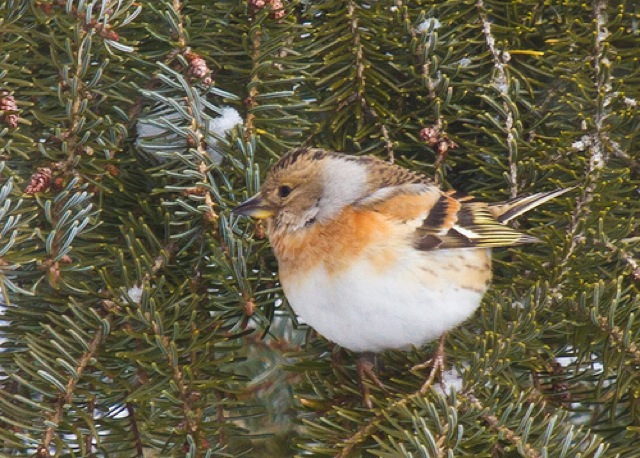}
  \end{subfigure}
  \begin{subfigure}[t]{\widthfactor\textwidth}
    \includegraphics[width=\imagewidth]{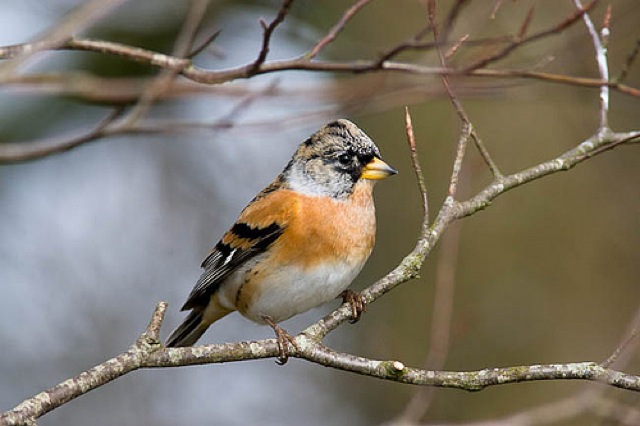}
  \end{subfigure}
  \begin{subfigure}[t]{\widthfactor\textwidth}
    \includegraphics[width=\imagewidth]{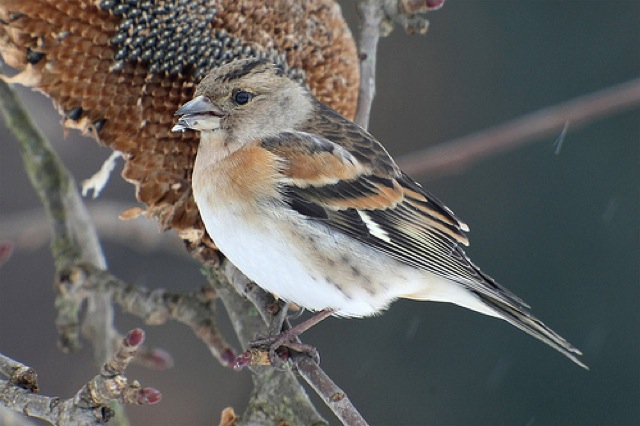}
  \end{subfigure} 
  \begin{subfigure}[t]{\widthfactor\textwidth}
    \includegraphics[width=\imagewidth]{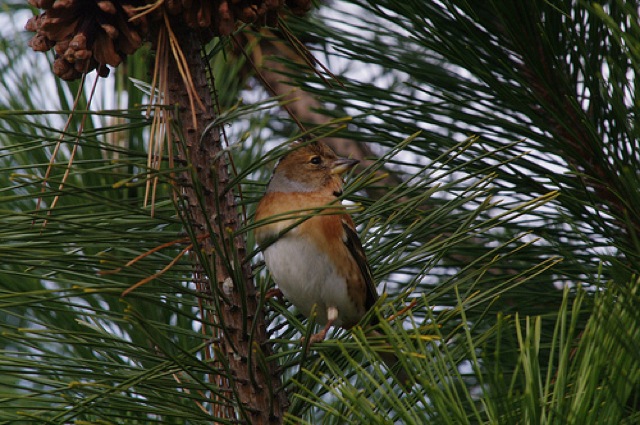}
  \end{subfigure}
  \begin{subfigure}[t]{\widthfactor\textwidth}
    \includegraphics[width=\imagewidth]{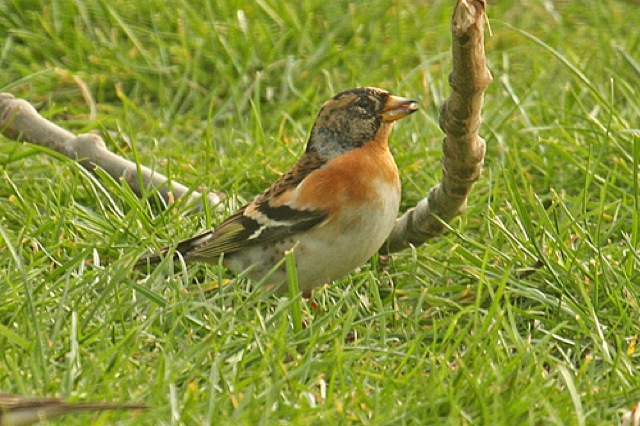}
  \end{subfigure}
  \begin{subfigure}[t]{\widthfactor\textwidth}
    \includegraphics[width=\imagewidth]{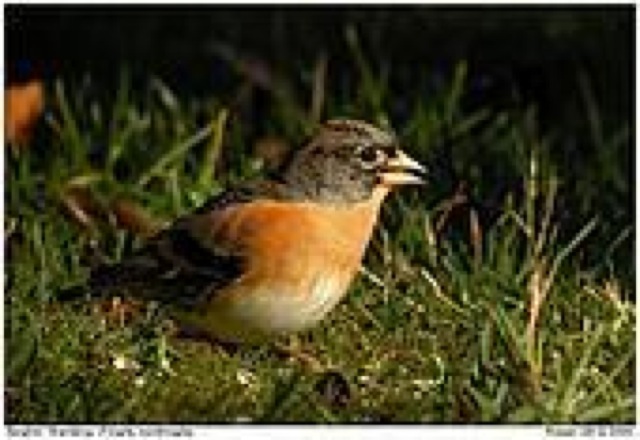}
  \end{subfigure}
  \\
  \begin{subfigure}[t]{\widthfactor\textwidth}
    \includegraphics[trim=0 4.7cm 0 3.0cm, clip=True, width=\imagewidth]{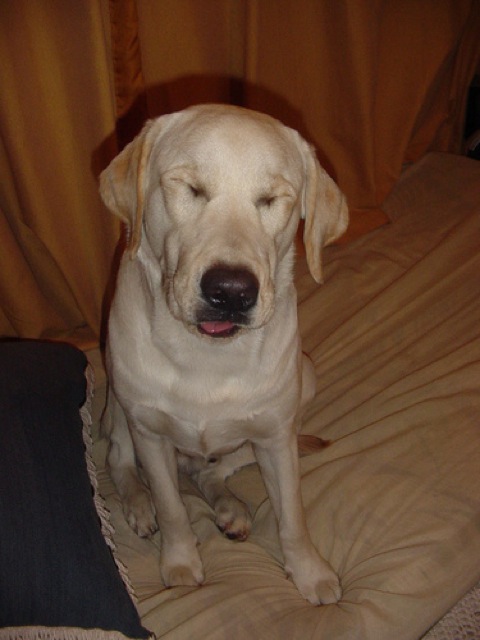}
  \end{subfigure}
  \begin{subfigure}[t]{\widthfactor\textwidth}
    \includegraphics[width=\imagewidth]{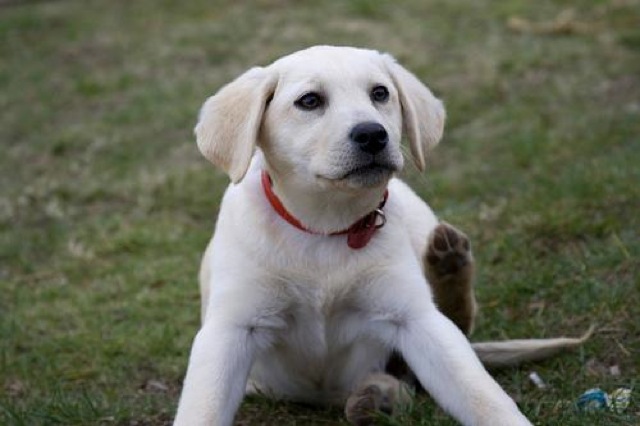}
  \end{subfigure}
  \begin{subfigure}[t]{\widthfactor\textwidth}
    \includegraphics[width=\imagewidth]{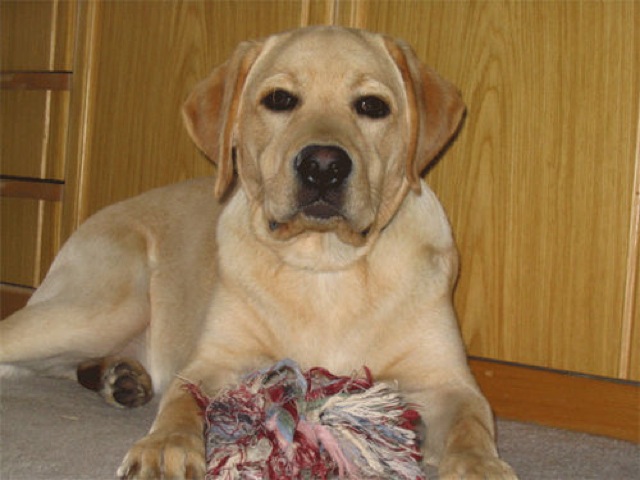}
  \end{subfigure} 
  \begin{subfigure}[t]{\widthfactor\textwidth}
    \includegraphics[width=\imagewidth]{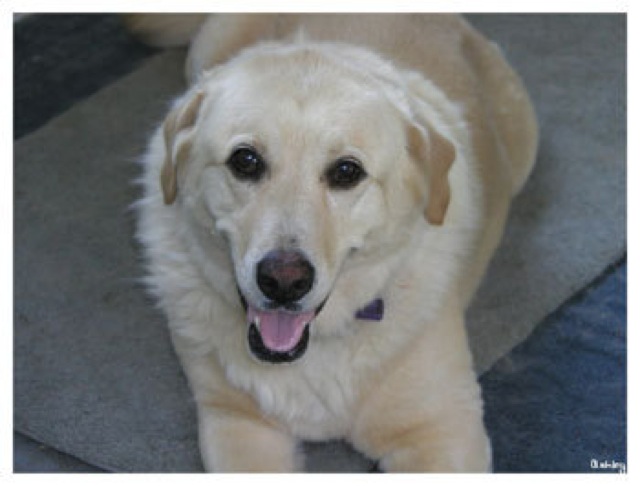}
  \end{subfigure}
  \begin{subfigure}[t]{\widthfactor\textwidth}
    \includegraphics[trim=0 2.5cm 0 2.5cm, clip=True, width=\imagewidth]{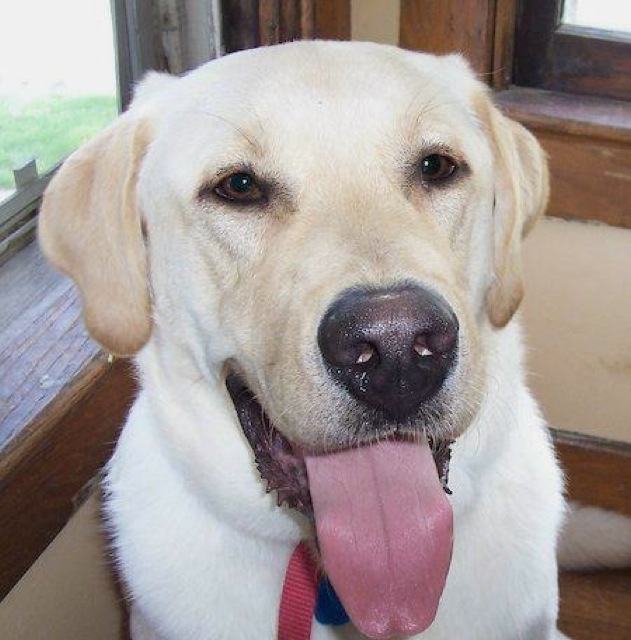}
  \end{subfigure}
  \begin{subfigure}[t]{\widthfactor\textwidth}
    \includegraphics[width=\imagewidth]{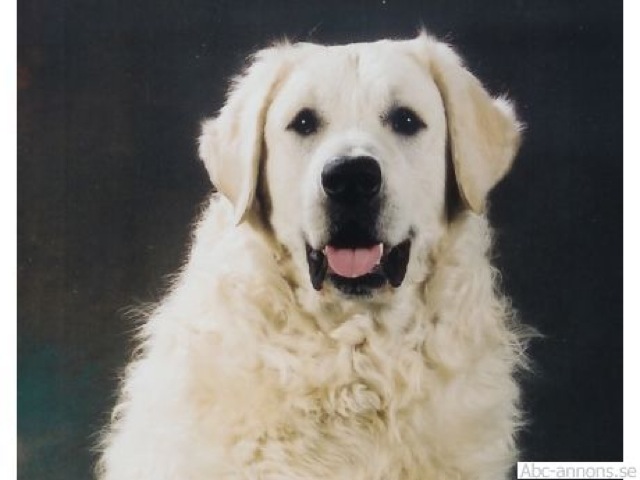}
  \end{subfigure}
  \\
  \begin{subfigure}[t]{\widthfactor\textwidth}
    \includegraphics[trim=0 2.5cm 0 2.5cm, clip=True, width=\imagewidth]{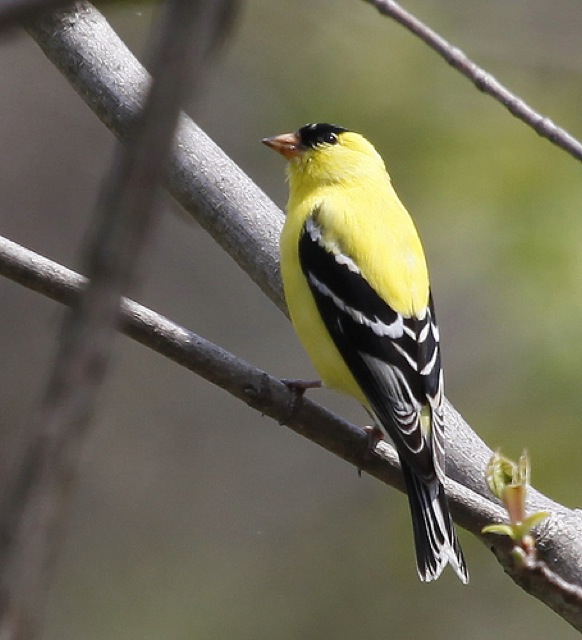}
  \end{subfigure}
  \begin{subfigure}[t]{\widthfactor\textwidth}
    \includegraphics[trim=0 1.5cm 0 1.5cm, clip=True, width=\imagewidth]{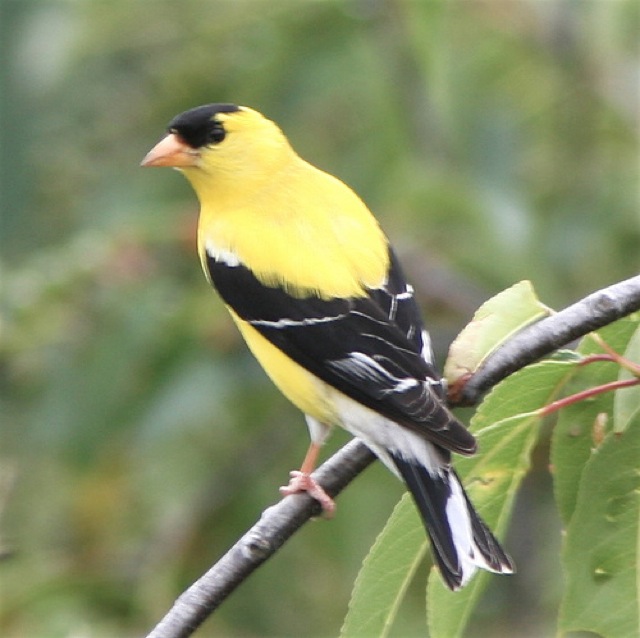}
  \end{subfigure}
  \begin{subfigure}[t]{\widthfactor\textwidth}
    \includegraphics[width=\imagewidth]{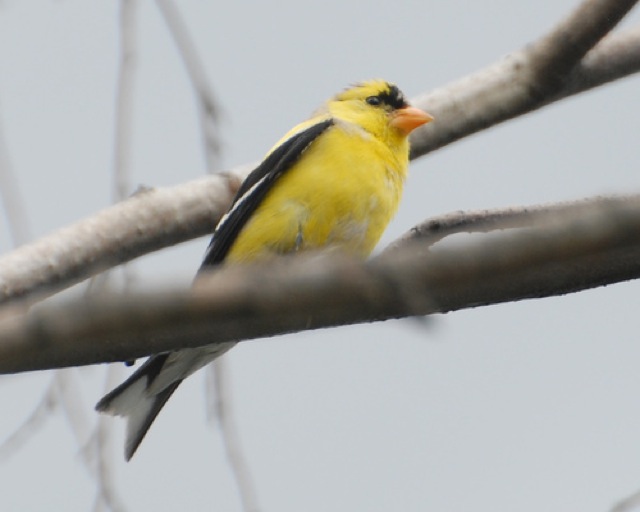}
  \end{subfigure} 
  \begin{subfigure}[t]{\widthfactor\textwidth}
    \includegraphics[width=\imagewidth]{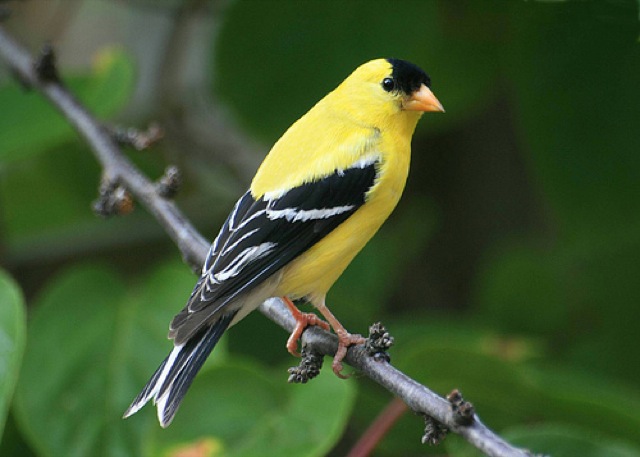}
  \end{subfigure}
  \begin{subfigure}[t]{\widthfactor\textwidth}
    \includegraphics[trim=0 0 0 2.5cm, clip=True, width=\imagewidth]{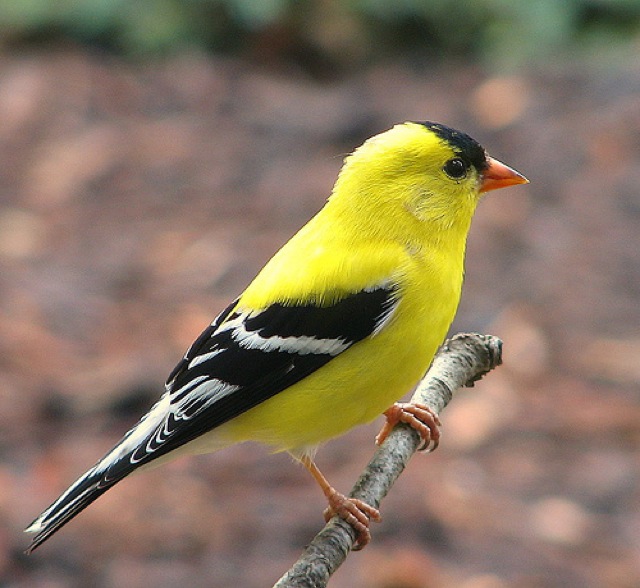}
  \end{subfigure}
  \begin{subfigure}[t]{\widthfactor\textwidth}
    \includegraphics[trim=0 2.5cm 0 2.5cm, clip=True, width=\imagewidth]{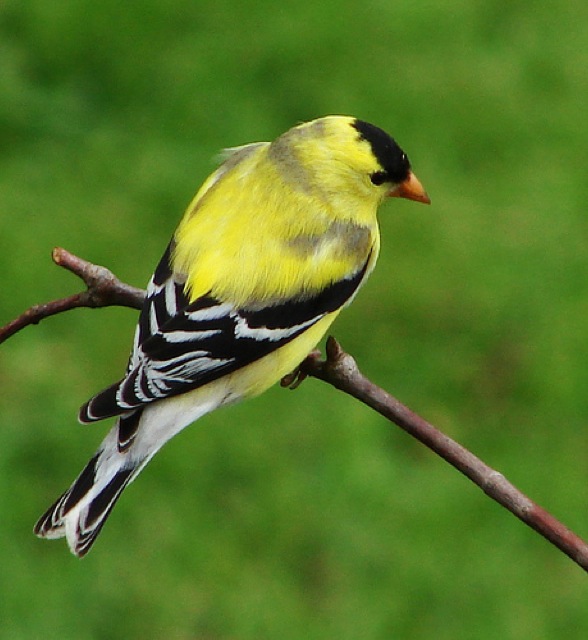}
  \end{subfigure}
  \\
  \begin{subfigure}[t]{\widthfactor\textwidth}
    \includegraphics[width=\imagewidth]{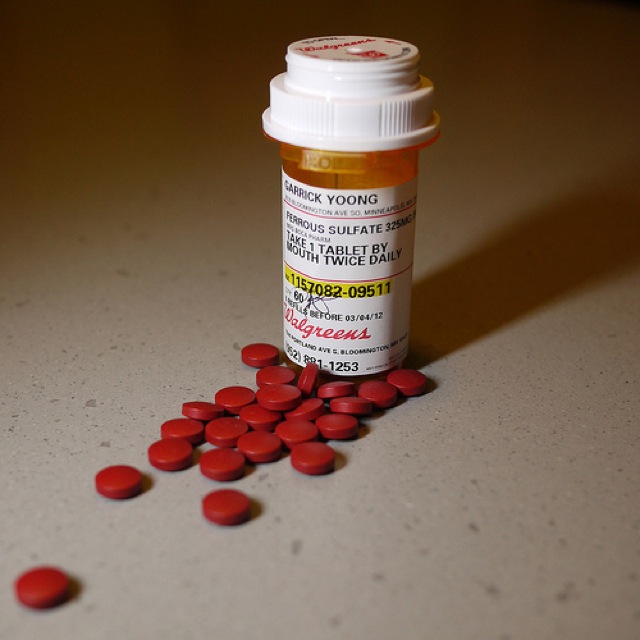}
  \end{subfigure}
  \begin{subfigure}[t]{\widthfactor\textwidth}
    \includegraphics[width=\imagewidth]{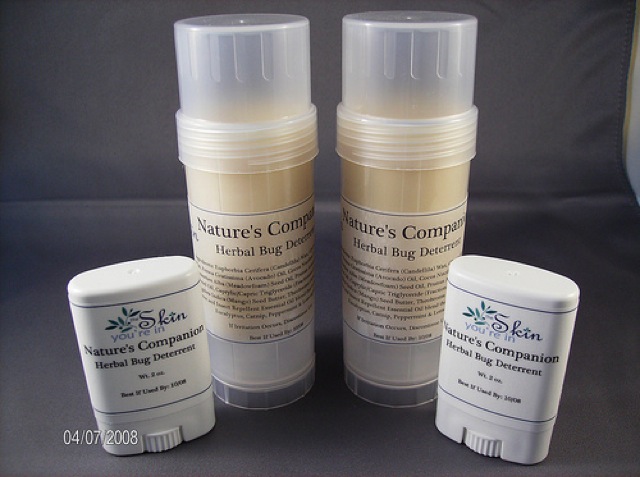}
  \end{subfigure}
  \begin{subfigure}[t]{\widthfactor\textwidth}
    \includegraphics[trim=0 4.5cm 0 1cm, clip=True, width=\imagewidth]{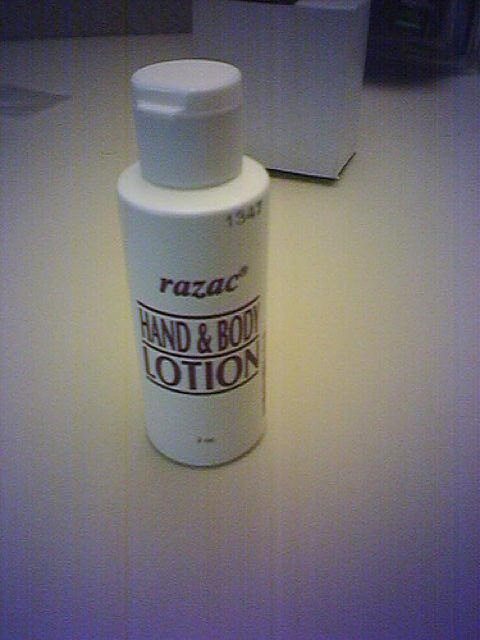}
  \end{subfigure} 
  \begin{subfigure}[t]{\widthfactor\textwidth}
    \includegraphics[width=\imagewidth]{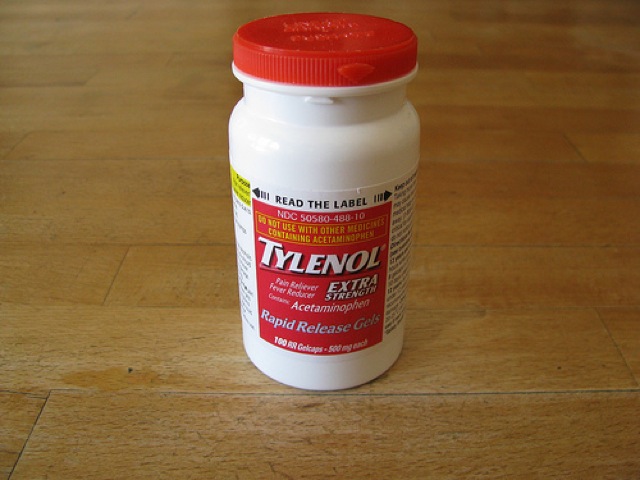}
  \end{subfigure}
  \begin{subfigure}[t]{\widthfactor\textwidth}
    \includegraphics[width=\imagewidth]{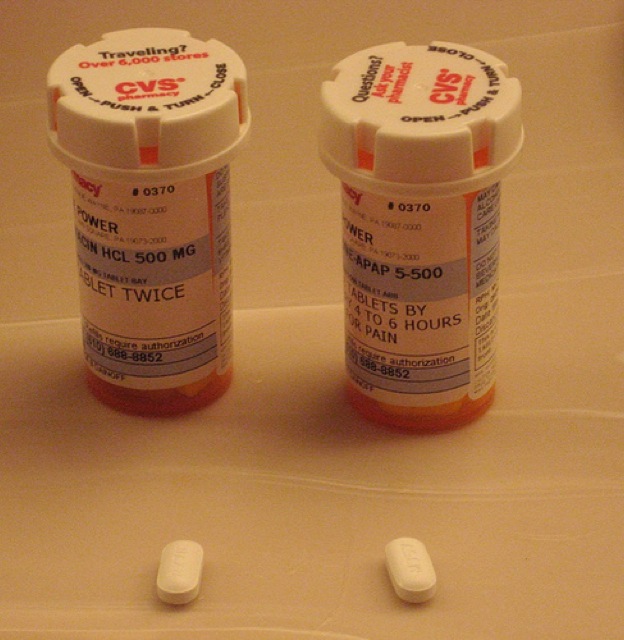}
  \end{subfigure}
  \begin{subfigure}[t]{\widthfactor\textwidth}
    \includegraphics[width=\imagewidth]{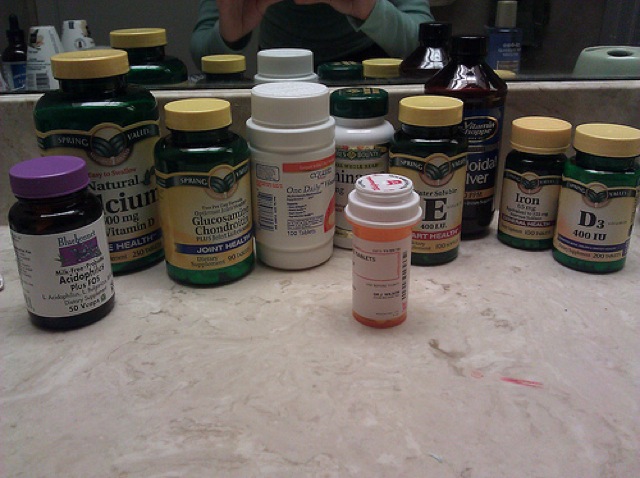}
  \end{subfigure}
  \caption{Retrieval based on subnetworks on the ILSVRC-2012 dataset. The 
  first image in each row is the query image; the remaining 5 are the 
  responses retrieved using submasks.}
  \label{fig:retrieval}
\end{figure}

We compare to DiffHash, a supervised similarity-preserving hashing approach proposed
by \citet{Strecha12}, trained on the non-binarized features from the network.
Supervision is represented in terms of similar and dissimilar pairs of points,
for which a ground-truth similarity measure is known, i.e. sharing the same
class or not. 
While it is beyond the scope of this paper to provide an exhaustive comparison or to
introduce a new approach to supervised hashing, we nevertheless show very
competitive performance w.r.t. a dedicated algorithm devised for this task.
Precision-recall curves are shown in Figure~\ref{fig:prcurves} while Table~\ref{tab:retrieval-table} 
reports results for mean average precision; $mAP = \sum_{r=1}^R P(r) \cdot
rel(r)$, where $rel(r)$ indicates the relevance of a result at a given rank
$r$, $P(r)$ the precision at $r$, and $R$ the number of retrieved results. 
DiffHash learns a linear projection, which is one of the reason we decided to use it 
to limit impact of supervision. Thus we attribute the small performance gap to the input 
features already being very discriminative which left little room for improvement.
For the purpose of this comparison, we did not investigate more sophisticated 
techniques which would have steered the focus to conventional hashing approaches.
Sample retrieval results for examples from the ILSVRC-2012
dataset are shown in Figure \ref{fig:retrieval}.

\section{Discussion} 

Training a system of many networks on a dataset such that they specialize 
to solve simpler tasks can be quite difficult without combining them into a 
single network with locally competitive units.
Without such local competition, one needs to have a global gating mechanism as
in \citet{jacobs1991}. The training algorithm and the objective function 
also need modifications such that competition between networks is encouraged,
and the system becomes hard to train. 
On the other hand, a locally competitive neural network can behave like a model
composed of many subnetworks, and massive sharing of parameters between
subnetworks enables better training. Stochastic gradient descent 
can be used to minimize the desired loss function,
and the implementation is so simple that one 
does not even realize that a model of models is being trained.

Figure \ref{fig:relu-mean-over-time} suggests that during optimization, 
the subnetworks get organized during an early transient phase such that 
subnetworks responding to similar examples have more parameters 
in common than those responding to dissimilar examples. 
This allows for better training of subnetworks due to reduced
interference from dissimilar examples and shared parameters for similar
examples. In the later fine-tuning phase, the parameters of subnetworks get
adjusted to improve classification and much less re-assignment of 
subnetworks is needed. 
In this way, the gating mechanism induced by locally competitive
activation functions accomplishes the purpose of global competition efficiently
and no modifications to the error function are required.

We believe that due to above advantages locally competitive networks
have allowed easier and faster training on complex pattern recognition tasks
compared to networks with sigmoidal or similar activation functions. 
These findings provide indirect evidence that low interference between 
subnetworks is a beneficial property for training large networks.
The nature of organization of subnetworks is reminiscent of the 
data manifold hypothesis for classification \citep{rifai2011}. 
Just like data points of different classes are expected to concentrate along 
sub-manifolds, we expect that the organization of subnetworks that respond to the 
data points reflects the data manifold being modeled. 

An important take-away from these results is the unifying theme between locally 
competitive architectures, which is related to past work on competitive learning. 
Insights from past literature on this topic may be utilized to develop improved 
learning algorithms and neural architectures.
This paper, to the best of our knowledge, is the first to show that useful binary data descriptors
can be obtained directly from a neural network trained for classification \emph{without any additional training}.
These descriptors are not just results of a thresholding trick or unique to a 
particular activation function, but arise as a direct result of the credit assignment process.
Our experiments on datasets of increasing complexity show that when the 
network performance improves, the performance gap 
to submask-based classification closes. This suggests that in the near future, as 
training techniques continue to advance and yield lower errors on larger 
datasets, submasks will perform as well as activation values for retrieval and 
transfer learning tasks. Importantly, these binary representations will always be 
far more efficient for storage and retrieval than continuous activation vectors.

\small
\bibliographystyle{iclr2015}
\bibliography{allrefs.bib}

\pagebreak
\appendix
\section{Supplementary Materials}
\subsection{Extra Visualizations}\label{sec:more-vis}
\begin{figure}[h]
  \centering 
  \begin{subfigure}[t]{0.49\textwidth}
  \centering 
    \includegraphics[scale=0.42]{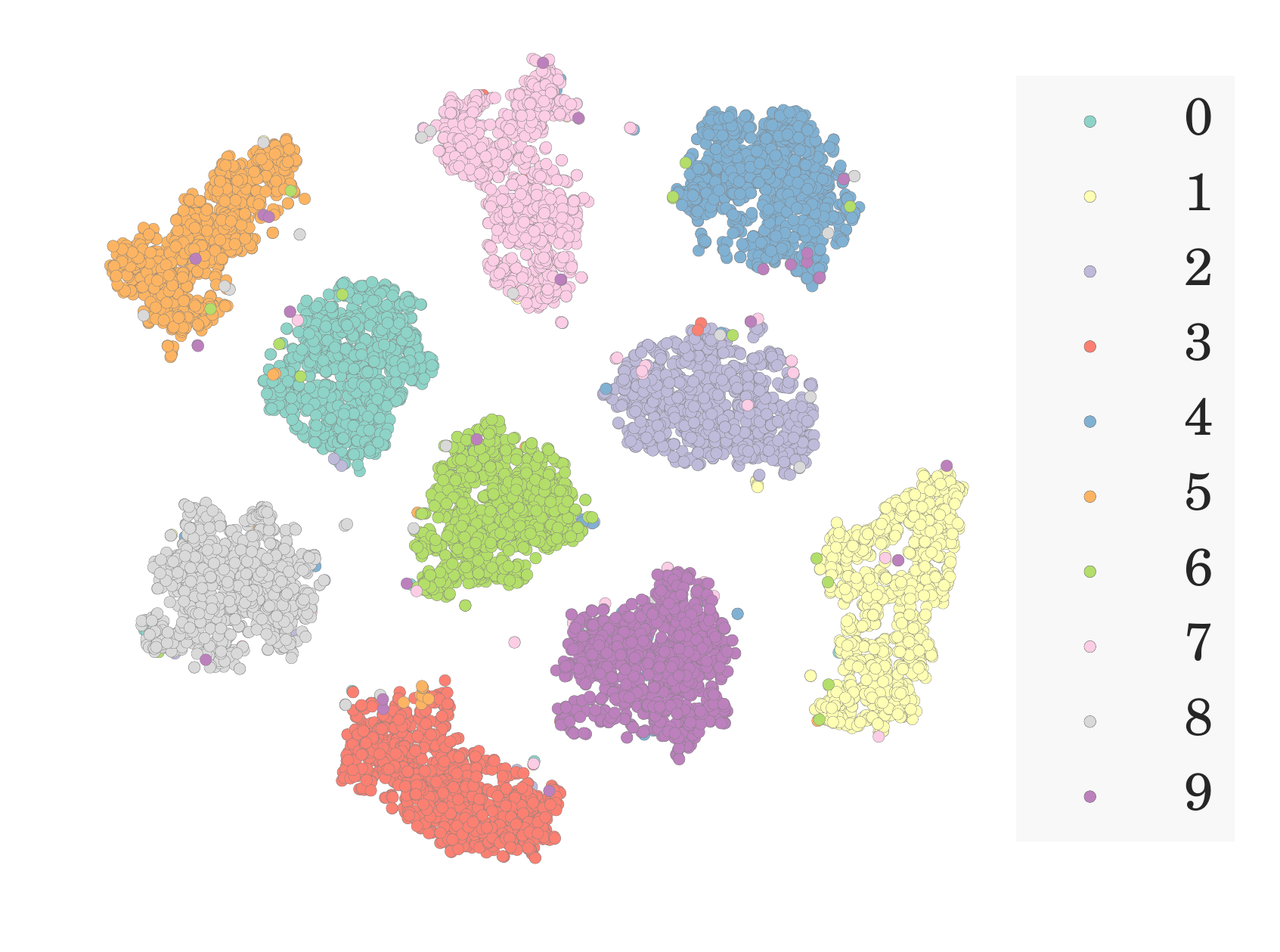} 
    \caption{Trained LWTA layer.}
    \label{fig:tsne-layer3-mnist-lwta-trained} 
  \end{subfigure} 
  \hfill
  \begin{subfigure}[t]{0.49\textwidth}
  \centering 
    \includegraphics[scale=0.42]{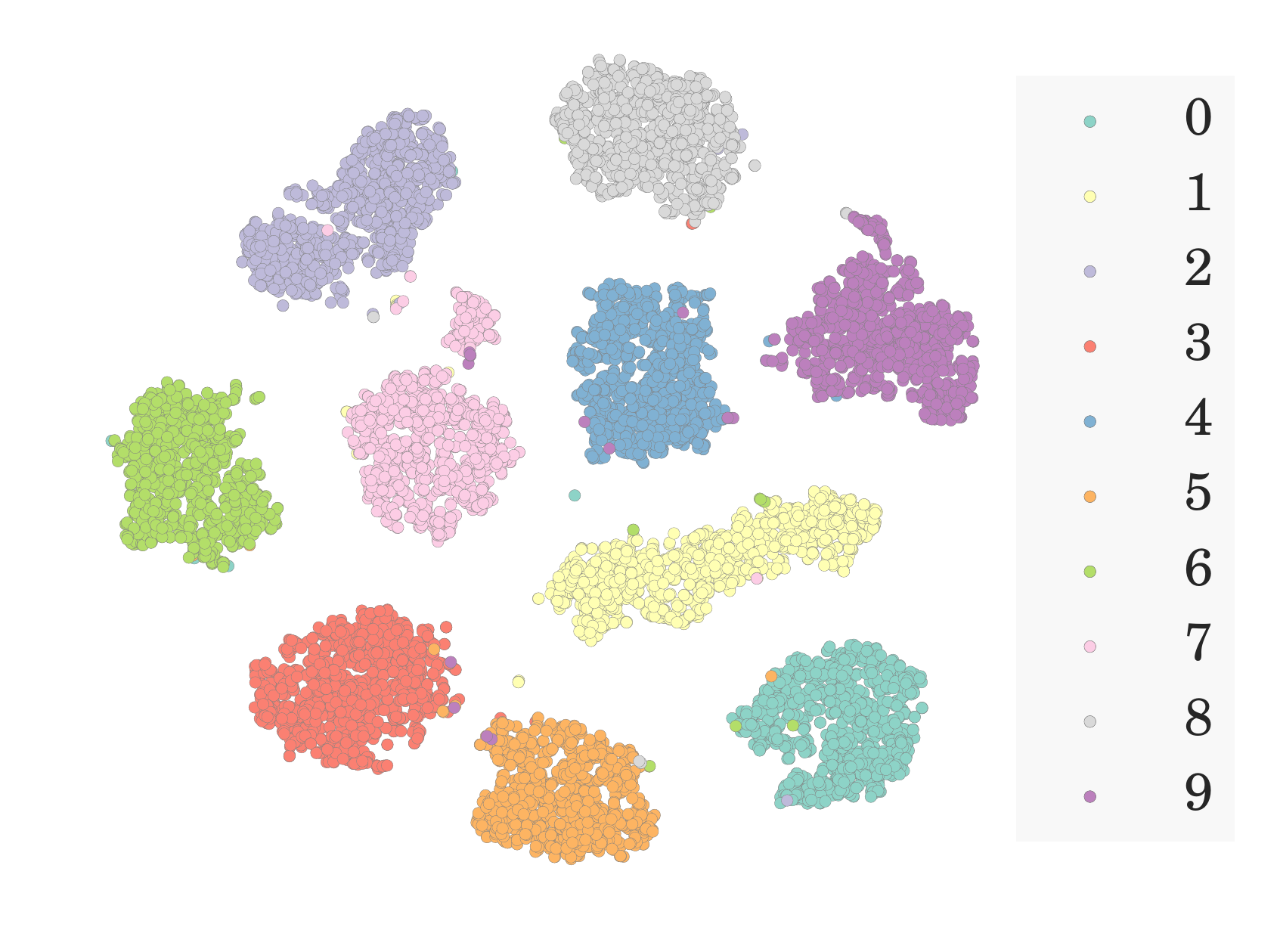} 
    \caption{Trained Maxout layer.}
    \label{fig:tsne-layer3-mnist-maxout-trained} 
  \end{subfigure} 
  \caption{2-D visualization of submasks from the penultimate layer
    of 3 hidden layer LWTA and maxout networks on MNIST test set. 
    Organization of submasks into distinct class specific clusters 
  similar to ReL networks is observed.}
  \label{fig:tsne-layer3-lwta-maxout}
\end{figure}

\begin{figure}[h]
  \centering 
  \begin{subfigure}[t]{0.49\textwidth}
  \centering 
    \includegraphics[scale=0.42]{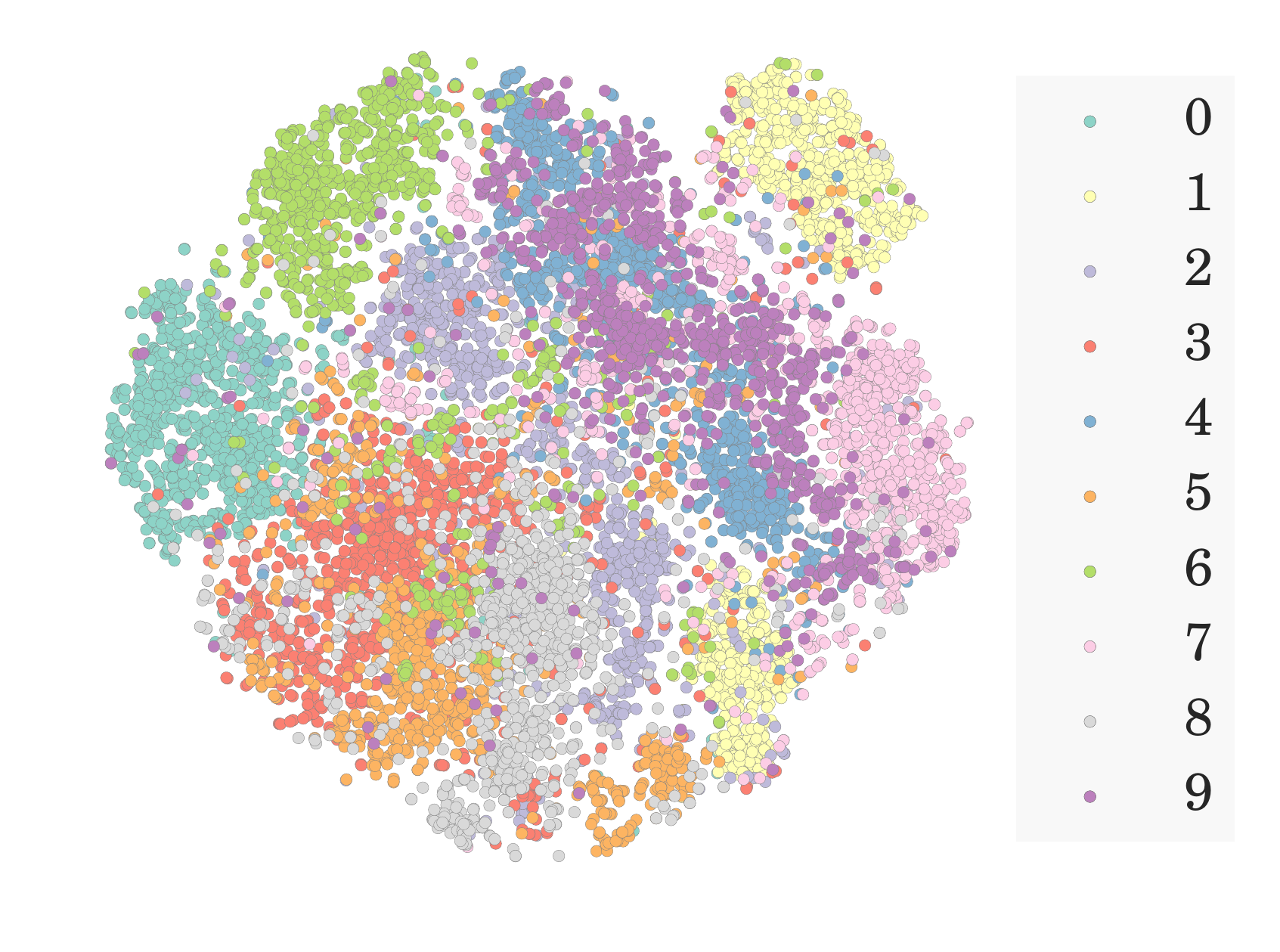} 
    \caption{Untrained 1st LWTA layer.}
    \label{fig:tsne-layer1-mnist-lwta-untrained} 
  \end{subfigure} 
  \hfill
  \begin{subfigure}[t]{0.49\textwidth}
  \centering 
    \includegraphics[scale=0.42]{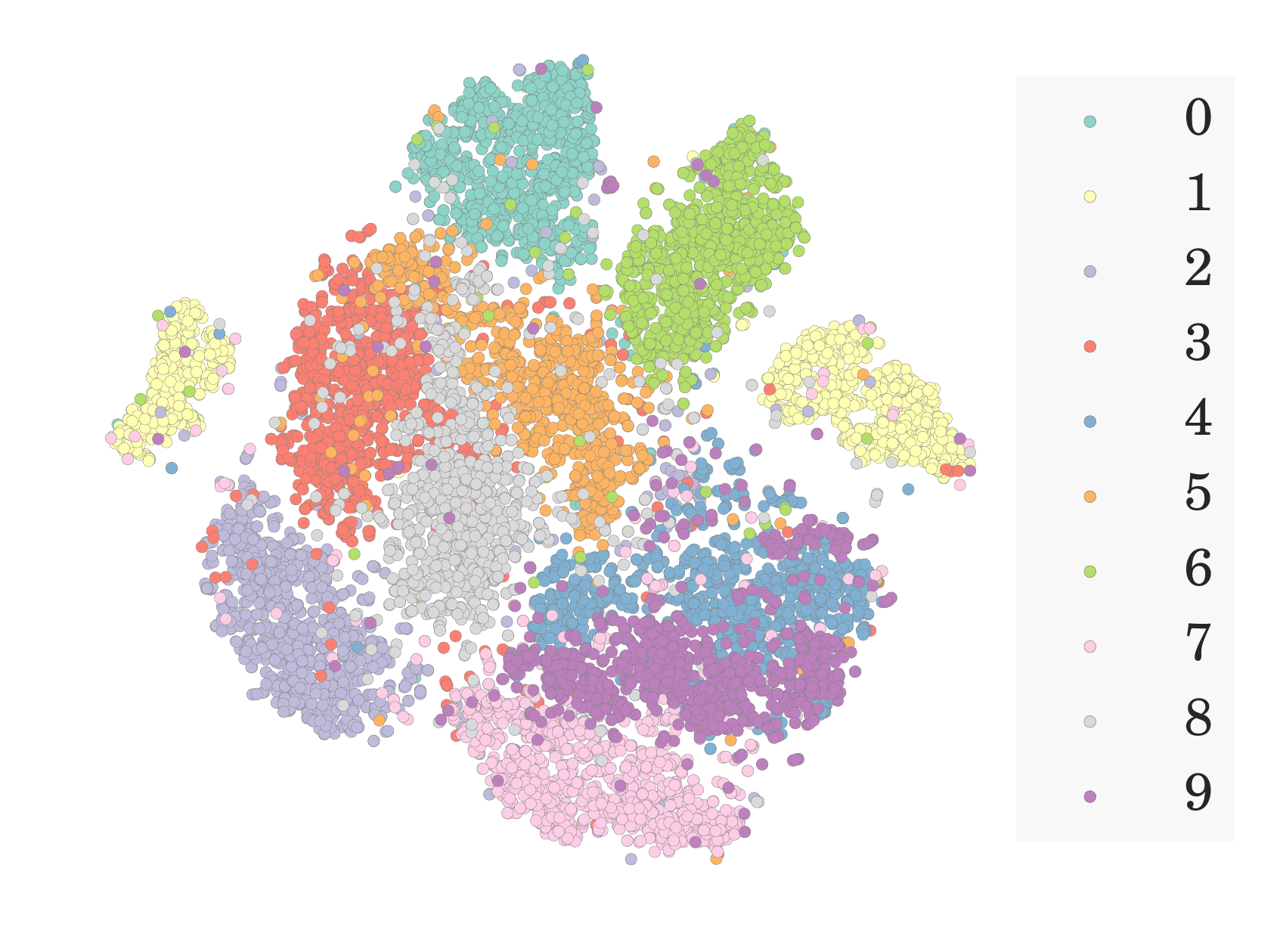} 
    \caption{Untrained 1st ReL layer.}
    \label{fig:tsne-layer1-mnist-relu-untrained} 
  \end{subfigure} 
  \caption{2-D visualization of submasks obtained before training
    from the 1st (closest to the input) hidden layer of 3 hidden layer 
  LWTA and ReL networks on MNIST test set.}
  \label{fig:tsne-layer1-lwta-maxout}
\end{figure}

\section{Dataset Descriptions}\label{sec:datasets}

\subsection{CIFAR-10 and CIFAR-100}
CIFAR-10 is a dataset of 32$\times$32 color images of 10 classes split into a training set 
of size 50,000 and testing set of size 10,000 (6000 images per class) 
\citep{krizhevsky2009}. CIFAR-100 is a similar dataset of color images 
but with 100 classes and 600 images per class, making it more challenging.
The models from \citet{goodfellow2013a} for these dataset utilize preprocessing 
using global contrast normalization and ZCA whitening as well as 
data augmentation using translational and horizontal reflections. 

\subsection{ImageNet (ILSVRC-2012)}
ILSVRC-2012 is a dataset of over a million natural images split into
1000 classes. An implementation of the network
in~\citet{krizhevsky2012}, with some minor
differences~\citep{donahue2013}, is available publicly.  For the
experiments in this section, the penultimate-layer activations
obtained using this model were downloaded from
CloudCV~\citet{batra2013}. The activations were obtained using the
center-only option, meaning that only the activations for the central,
224$\times$224 crop of each image were used. 

For each validation set example, 100 examples from the 
training set with the closest submasks were weighted by
the inverse of the distance, then the classes with top-1 or 
top-5 weighted sums were returned as predictions. 

\section{Note on Sigmoidal Networks}

In this paper we focused on improving our understanding of neural networks
with locally competitive activation functions. We also
obtained binary codes for efficient retrieval directly from neural networks
trained for classification, but this was not the primary aim of our study. 
When this is the aim, we note here that it possible to use sigmoidal 
activation functions to obtain binary codes by thresholding the
activation values after supervised or unsupervised \citep{salakhutdinov2009} 
training. However it should be noted that:

\begin{itemize}
\item
The thresholding is somewhat arbitrary and the best threshold needs to be selected
by trying various values. For locally competitive networks, the binarization is natural
and inherent to the nature of credit assignment in these networks.

\item
Since sigmoidal networks are hard and slow to train, the approach of thresholding their 
activations is impractical for large datasets which are common for retrieval tasks. 
Locally competitive networks have been crucial for the successful application of
neural networks to such datasets.
\end{itemize}
\end{document}